\documentclass[journal]{IEEEtran}
\usepackage[caption=false,font=normalsize,labelfont=sf,textfont=sf]{subfig}
\usepackage{stfloats}
\usepackage{verbatim}
\usepackage{cite}
\usepackage{graphicx} 
\usepackage{multirow} 
\usepackage{amsmath, amssymb, amsfonts} 
\usepackage{amsthm} 
\usepackage{mathrsfs} 
\usepackage{balance}
\usepackage{xcolor} 
\usepackage{textcomp} 
\usepackage{booktabs} 
\usepackage{algorithm} 
\usepackage{algpseudocode} %
\usepackage{listings} 
\usepackage{array} 
\usepackage{tabularx} 
\usepackage{url} 
\usepackage{makecell} 
\usepackage[hidelinks,hypertexnames=false,colorlinks=true,citecolor=red!70!black,urlcolor=black]{hyperref}
\usepackage{caption}
\usepackage{orcidlink}
\newcommand{\settablefont}{\fontsize{7}{11.0}\selectfont}
\usepackage{tikz}
\newcommand\normx[1]{\left\Vert#1\right\Vert}
\newcolumntype{L}[1]{>{\raggedright\arraybackslash}p{#1}}
\newcolumntype{C}[1]{>{\centering\arraybackslash}p{#1}}
\newcolumntype{R}[1]{>{\raggedleft\arraybackslash}p{#1}}

\newcommand{\ie}{\textit{i.e.}}

\newcommand{{\etal}}{\textit{et al.}}

\definecolor{SC-Depth}{HTML}{2E2E2E}
\definecolor{SCIPaD}{HTML}{8B4513}
\definecolor{Nist}{HTML}{00B050}
\definecolor{Helmke}{HTML}{CCAA33}
\definecolor{zhao2024}{HTML}{0000FF}
\definecolor{zhao2020}{HTML}{29BDFF}
\definecolor{ling}{HTML}{8E44AD}
\definecolor{SC-Depth}{HTML}{2E2E2E}
\definecolor{MyPink}{HTML}{e377c2}
\definecolor{MyOrange}{HTML}{ff7f0e}
\definecolor{MyGreen}{HTML}{2ca02c}

\begin{document}

\title{A Birotation Solution for Relative Pose Problems}

\author{
    Hongbo~Zhao\,\orcidlink{0009-0008-2198-2484}
    Ziwei~Long\,\orcidlink{0009-0007-1520-5526} 
    Mengtan~Zhang\,\orcidlink{0009-0003-3468-7680}
    Hanli~Wang\,\orcidlink{0000-0002-9999-4871},~\IEEEmembership{Senior Member,~IEEE}, \\
    Qijun~Chen\,\orcidlink{0000-0001-5644-1188},~\IEEEmembership{Senior Member,~IEEE},
    and Rui~Fan\,\orcidlink{0000-0003-2593-6596},~\IEEEmembership{Senior Member,~IEEE}%
    \thanks{\textit{Corresponding author: Rui Fan}.}
    \thanks{Hongbo Zhao and Mengtan Zhang are with the Shanghai Research Institute for Intelligent Autonomous Systems, Tongji University, Shanghai, P. R. China (e-mail: hongbozhao@tongji.edu.cn; zmt200110@tongji.edu.cn).}%
    \thanks{Ziwei Long and Qijun Chen are with the College of Electronics \& Information Engineering, Tongji University, Shanghai, P. R. China (e-mail: zwlong@tongji.edu.cn; qjchen@tongji.edu.cn).}%
    \thanks{Hanli Wang is with the College of Electronics and Information Engineering, Tongji University, Shanghai, P. R. China, the School of Computer Science and Technology, Tongji University, Shanghai, P. R. China, and the Key Laboratory of Embedded System and Service Computing (Ministry of Education), Tongji University, Shanghai, P. R. China (e-mail: hanliwang@tongji.edu.cn).}
    \thanks{
    Rui Fan is with the College of Electronics \& Information Engineering, Shanghai Institute of Intelligent Science and Technology, Shanghai Research Institute for Intelligent Autonomous Systems, the State Key Laboratory of Intelligent Autonomous Systems, and Frontiers Science Center for Intelligent Autonomous Systems, Tongji University, Shanghai, P. R. China (e-mail: rui.fan@ieee.org).}
}
\maketitle
\begin{abstract}
    Relative pose estimation, a fundamental computer vision problem, has been extensively studied for decades. Existing methods either estimate and decompose the essential matrix or directly estimate the rotation and translation to obtain the solution. In this article, we break the mold by tackling this traditional problem with a novel birotation solution. We first introduce three basis transformations, each associated with a geometric metric to quantify the distance between the relative pose to be estimated and its corresponding basis transformation. Three energy functions, designed based on these metrics, are then minimized on the Riemannian manifold $\mathrm{SO(3)}$ by iteratively updating the two rotation matrices. The two rotation matrices and the basis transformation corresponding to the minimum energy are ultimately utilized to recover the relative pose. Extensive quantitative and qualitative evaluations across diverse relative pose estimation tasks demonstrate the superior performance of our proposed birotation solution. Source code, demo video, and datasets will be available at \href{https://mias.group/birotation-solution}{mias.group/birotation-solution} upon publication.
\end{abstract}

\section{Introduction}
\label{sec.intro}

\IEEEPARstart{R}{elative} pose estimation is the core of various computer vision applications, including stereo camera extrinsic calibration \cite{dexheimer2022information, dang2009continuous, hansen2012online}, visual odometry \cite{nister2004visual, levin2004visual, chen2024leap}, and so forth. This process estimates a transformation matrix, defined as follows:
\begin{equation}
\boldsymbol{P}=
    \begin{pmatrix}
        \boldsymbol{R} & \boldsymbol{t}\\
        \boldsymbol{0}^\top & 1
    \end{pmatrix},
    \label{eq.transformation_matrix}
\end{equation}
where $\boldsymbol{R} \in \mathrm{SO(3)}$ denotes the rotation matrix, $\boldsymbol{t}\in\mathbb{R}^3$ represents the translation vector, and $\boldsymbol{0}$ is a column vector of zeros. This problem can be solved by estimating an essential matrix $\boldsymbol{E}$, which satisfies the following epipolar constraint:
\begin{equation}
{\boldsymbol{p}^C_2}^{\top}\boldsymbol{E}{\boldsymbol{p}^C_1}=0,
    \label{eq.plEpr}
\end{equation}
and decomposing it into $\boldsymbol{R}$ and $\boldsymbol{t}$,
where $\boldsymbol{p}^C_1=(x_1, y_1, z_1)^{\top}$ and $\boldsymbol{p}^C_2=(x_2, y_2, z_2)^{\top}$ represent a pair of 3-D correspondences in the reference and target camera coordinate systems, respectively\footnote{In this article, the subscripts ``1" and ``2" denote the reference and target views, respectively.
}. Alternatively, $\boldsymbol{P}$ can be directly estimated based on the following relation:
\begin{equation}
\tilde{\boldsymbol{p}}^C_2=\boldsymbol{P}\tilde{\boldsymbol{p}}^C_1,
    \label{eq.p12p2}
\end{equation}
where $\tilde{\boldsymbol{p}}^C$ denotes the homogeneous coordinates of $\boldsymbol{p}^C$.  Although both methods provide viable solutions for determining the relative pose, they still have significant limitations.

In the first approach, the essential matrix inherently possesses only five degrees of freedom (DoFs), yet its estimation typically requires solving for eight or nine unknown parameters, which can undermine both the uniqueness and stability of the solution \cite{hartley1997defense, nister2004efficient}. In particular, overfitting to inaccurately matched correspondences can substantially impair the performance of relative pose estimation \cite{chum2008optimal}. To address this issue, researchers have resorted to correspondence refinement strategies \cite{wei2023generalized, jau2020deep, zhao2020efficient, barath2019magsac} and additional geometric constraints \cite{nister2004efficient, helmke2007essential}. Additionally, this approach cannot solve pure rotation problems, as the essential matrix degenerates to a zero matrix under such conditions \cite{cai2019equivalent}.

In the second approach, the choice of the optimization algorithm hinges on the availability of depth information. When depth is known, relative pose estimation can be formulated as a perspective-n-point (PnP) problem \cite{zheng2013revisiting}, which can be solved reliably and efficiently using methods such as efficient PnP \cite{lepetit2009ep}. In contrast, when depth is unknown, the scale of the translation vector $\boldsymbol{t}$ becomes ambiguous, necessitating its treatment as a unit vector \cite{ling2016high}. Normalizing $\boldsymbol{t}$, nevertheless, increases the risk of the optimization process falling into local minima, posing significant challenges in achieving stable and accurate solutions \cite{zhao2024dive}.

Substituting the translation vector with a rotation matrix and using two rotation matrices to solve the relative pose problem can avoid the instability caused by the absence of depth information while introducing only one additional dimensionality to the solution space \cite{zhao2024dive}. Although this concept has been investigated in previous studies \cite{dang2009continuous, hansen2012online, zhao2024dive}, its practical application remains predominantly confined to stereo camera extrinsic online calibration, where the rotation matrix approximates the identity matrix and the translation vector aligns closely with the X-axis \cite{hansen2012online, zhao2024dive}. Therefore, these methods fall short of providing a stable solution for general relative pose estimation problems.

To address this well-studied yet enduring problem, we propose a novel birotation solution in this article. Our study effectively expands existing methods, initially designed for stereo camera extrinsic online calibration, to tackle a broader range of relative pose problems. Specifically, we first define three basis transformations along the X-, Y-, and Z-axes. Three geometric metrics are then designed to quantify the distances between the relative pose to be estimated and these basis transformations. Each metric can be employed to construct an energy function $E_i$, which includes a regularization term to guarantee the uniqueness of the solution. The energy function $E_i$ is minimized on the Riemannian manifold $\mathrm{SO(3)}$ through iterative updates of the reference and target camera coordinate systems with two independent rotation matrices, $\boldsymbol{R}_{1, i}$ and $\boldsymbol{R}_{2, i}$. These optimization processes yield three locally optimal birotation solutions, each defined by a pair of rotations, a scaling factor, and a coordinate axis. The birotation solution with the minimal geometric metric is then selected as the optimal solution, from which the relative pose $\boldsymbol{P}$ can be directly derived. Extensive experiments across a variety of relative pose estimation tasks demonstrate the feasibility of our proposed solution and its greater accuracy over all other state-of-the-art (SoTA) methods.

In a nutshell, our contributions are as follows:
\begin{itemize}
    \item We revisit the relative pose problem from a fresh perspective and propose a novel birotation solution.
    \item We introduce three basis transformations and corresponding geometric metrics to expand relevant prior works, which primarily focus on stereo camera extrinsic calibration, to solve general relative pose problems.
    \item We propose a highly efficient algorithm for relative pose estimation based on the birotation solution, which simultaneously optimizes three energy functions derived from geometric metrics.
    \item We demonstrate the feasibility and superiority of our solution through quantitative and qualitative evaluations across diverse relative pose estimation tasks.
\end{itemize}

The remainder of this article is organized as follows: Sect.~\ref{sec.relatedwork} reviews previous relative pose estimation algorithms. The proposed basis transformations, geometric metrics, birotation solutions, and the relative pose estimation algorithm are detailed in Sect.~\ref{sec.method}. Sect.~\ref{sec.Experiments} presents experimental results that demonstrate the effectiveness and superiority of our algorithm. The potential risks associated with the uniqueness of the birotation solution are discussed in Sect.~\ref{sec.discussion}. Finally, we conclude this article and suggest directions for future work in Sect.~\ref{sec.Conclusion}.

\section{Related Work}
\label{sec.relatedwork}

For the relative pose problem, the essential matrix estimation process under the epipolar constraint in \eqref{eq.plEpr} can be formulated as follows:
\begin{equation}
    \underset{\boldsymbol{E}}{\arg\min}\sum^{N}_{n = 1}\| {\boldsymbol{p}^C_{2,n}}^{\top}\boldsymbol{E}{\boldsymbol{p}^C_{1,n}} \|^2_2,
    \label{eq.E_solution}
\end{equation}
where $N$ represents the number of 3-D correspondences, and $\boldsymbol{p}^{C}_{1, n}$ and $\boldsymbol{p}^{C}_{2, n}$ denote the $n$-th pair of 3-D correspondences in the reference and target camera coordinate systems, respectively. Although the eight-point method \cite{hartley1997defense} can effectively solve such an $N$-point problem, it necessitates highly accurate correspondences. The five-point method, combined with random sample consensus (RANSAC) \cite{nister2004efficient, chum2008optimal}, proves effective in addressing cases where correspondences are less reliable. Furthermore, existing local optimization approaches \cite{kanatani2010unified, zhang1998determining, helmke2007essential} either suffer from low efficiency or are prone to becoming trapped in local minima. 
While manifold-based approaches \cite{ma2001optimization, tron2017space, helmke2007essential} provide insightful constructions of the solution space, they often fail to guarantee iterative convergence in complex relative pose problems \cite{zhao2020efficient}.
To overcome these challenges, the study \cite{zhao2020efficient} reformulates the $N$-point problem as a quadratically constrained quadratic program (QCQP) and utilizes semi-definite relaxation to achieve a globally optimal solution, demonstrating polynomial-time convergence. 

Different from the above-mentioned methods, the study \cite{ling2016high} directly estimates $\boldsymbol{R}$ and $\boldsymbol{t}$ by solving the following optimization problem based on the epipolar constraint:
\begin{equation}
\underset{\boldsymbol{R},\boldsymbol{t}}{\arg\min} \sum^{N}_{n = 1}\normx{{\boldsymbol{p}^C_{1,n}}^\top [\boldsymbol{t}]_\times\boldsymbol{R} \boldsymbol{p}^C_{2,n}}^2_2,
    \label{eq.Rt_solution}
\end{equation} 
where $[\boldsymbol{t}]_{\times}$ denotes cross-product matrix of $\boldsymbol{t}$. However, this method demonstrates limited stability in estimating $\boldsymbol{t}$, primarily due to the significant non-convexity of the energy function introduced by $[\boldsymbol{t}]_{\times}$ and the normalization process of $\boldsymbol{t}$ \cite{zhao2024dive}. 

In recent years, data-driven methods, such as PoseNet \cite{kendall2015posenet}, have gained significant popularity. Prior studies predominantly utilize convolutional neural networks, often employing a ResNet-18, to directly regress the 6-DoF relative poses, which include three Euler angles and three translation vector elements. Nevertheless, these methods often demonstrate limited interpretability, generalization, and accuracy. Recent advancements, such as SC-Depth \cite{bian2021unsupervised} and SCIPaD \cite{feng2024scipad}, have incorporated a monocular depth estimation branch to improve the overall performance of relative pose estimation. However, the insufficient geometric constraints between depth and relative pose incorporated during training limit the accuracy of these approaches.

As discussed above, the use of birotation for relative pose estimation remains confined to stereo camera extrinsic rectification. The study \cite{hansen2012online} uses two independent rotation matrices, $\boldsymbol{R}_1$ and $\boldsymbol{R}_2$, to separately rotate the left and right camera coordinate systems. They derive the Euler angles for $\boldsymbol{R}_1$ and $\boldsymbol{R}_2$ by formulating and solving the following optimization problem based on stereo rectification:
\begin{equation}
    \underset{\boldsymbol{R}_1,\boldsymbol{R}_2}{\arg\min} \sum^{N}_{n=1} \normx{\frac{\boldsymbol{r}_{1,2}^{\top}\boldsymbol{p}^{C}_{1,n}}{\boldsymbol{r}_{1,3}^{\top}\boldsymbol{p}^{C}_{1,n}}-\frac{\boldsymbol{r}_{2,2}^{\top}\boldsymbol{p}^{C}_{2,n}}{\boldsymbol{r}_{2,3}^{\top}\boldsymbol{p}^{C}_{2,n}}}^2_2,
\end{equation}
where $\boldsymbol{r}_{1,j}$ and $\boldsymbol{r}_{2,j}$ denote the $j$-th row vector of $\boldsymbol{R}_1$ and $\boldsymbol{R}_2$, respectively. The recent study \cite{zhao2024dive} improves this approach by incorporating geometry constraints to ensure solution uniqueness. Nonetheless, these approaches are applicable only to relatively ideal stereo vision models (with small perturbations) and cannot be generalized for broader relative pose estimation problems.

\section{Methodology}
\label{sec.method}

While it is geometrically feasible to obtain an ideal stereo vision model (a transformation along the X-axis) by rotating two cameras, solving for such a pure rotation problem becomes highly challenging, particularly when the actual relative pose significantly deviates from the ideal model. This challenge arises because existing birotation-based methods \cite{hansen2012online, zhao2024dive} primarily utilize local optimization algorithms to address a non-convex problem. Their effectiveness is influenced by both the initialization and the penalty term in each energy function, which can misdirect the optimization process. This greatly limits their applicability to general relative pose problems. Therefore, Sect. \ref{sec.Basis_Transformations} first introduces two additional transformations along the Y- and Z-axes, respectively. Subsequently, Sect. \ref{sec.Geometric_Metrics} defines the geometric metrics corresponding to these three basis transformations. These geometric metrics are then employed to formulate energy functions, the minimization of which yields three birotation solutions, as presented in Sect. \ref{sec.The_Birotation_Solution}. Building upon these solutions, a novel relative pose estimation algorithm is introduced in Sect. \ref{sec.Relative_Pose_Estimation_Algorithm}.

\subsection{Basis Transformations}
\label{sec.Basis_Transformations}

Previous birotation-based methods \cite{hansen2012online, zhao2024dive} have been confined to an ideal stereo vision model (both cameras have an identical intrinsic matrix), primarily because their energy functions are formulated based on the constraint that $\boldsymbol{p}_1 = (u_1, v_1)^{\top}$ and $\boldsymbol{p}_2 = (u_2, v_2)^{\top}$, a given pair of correspondences in the two images, have the same vertical coordinates (\ie, $v_1=v_2$). Similar constraints can be observed when a single camera translates solely along either the Y-axis or the Z-axis. Specifically, the horizontal coordinates of $\boldsymbol{p}_1$ and $\boldsymbol{p}_2$ are identical (\ie, $u_1=u_2$) in the former case, while the ratios of their horizontal to vertical coordinates are identical (\ie, $\frac{u_1}{v_1}=\frac{u_2}{v_2}$) in the latter case.

Nevertheless, in general relative pose problems, the intrinsic matrices of the reference and target cameras may differ. It is therefore necessary to reformulate the above-mentioned constraints based on the following basis transformations:
\begin{equation}
    {\boldsymbol{p}^{C}_1} = {\boldsymbol{p}^{C}_2} + s_i\boldsymbol{l}_i, 
    \ \ i\in \{1, 2, 3\},
 \label{eq.Basis_Transformations}
\end{equation}
where $\boldsymbol{l}_1 = (1,0,0)^\top$, $\boldsymbol{l}_2 = (0,1,0)^\top$, and $\boldsymbol{l}_3 = (0,0,1)^\top$ are unit vectors representing pure translations along the X-, Y-, and Z-axes of the camera coordinate system, respectively, and $s_i \in (-\infty, +\infty)$ denotes the scaling factor corresponding to each transformation. $\boldsymbol{p}^C_{1,2}$ and $\boldsymbol{p}_{1,2}$ have the following relation:
\begin{equation}
\begin{cases}
\boldsymbol{p}^C_1 = z_1\bar{\boldsymbol{p}}_1 = z_1\boldsymbol{K}^{-1}_1\tilde{\boldsymbol{p}}_1, \\ 
\boldsymbol{p}^C_2 = z_2\bar{\boldsymbol{p}}_2 = z_2\boldsymbol{K}^{-1}_2\tilde{\boldsymbol{p}}_2,
\end{cases}
 \label{eq.camera_model}
\end{equation}
where $\bar{\boldsymbol{p}}_{1,2}$ represents the normalized coordinates of $\boldsymbol{p}^{C}_{1,2}$ with respect to depth, $\tilde{\boldsymbol{p}}_{1,2} = (u_{1,2}, v_{1,2}, 1)^\top$ denotes the homogeneous coordinates of $\boldsymbol{p}_{1,2}$, and 
\begin{equation}
    \boldsymbol{K}_1 = \begin{pmatrix}
        f_{x,1} & 0   & u_{o,1} \\
        0   & f_{y,1} & v_{o,1} \\
        0   & 0   & 1
    \end{pmatrix}
    \label{eq.K_1}
\end{equation}
and
\begin{equation}
    \boldsymbol{K}_2 = \begin{pmatrix}
        f_{x,2} & 0   & u_{o,2} \\
        0   & f_{y,2} & v_{o,2} \\
        0   & 0   & 1
    \end{pmatrix}
    \label{eq.K_2}
\end{equation}
represent the intrinsic matrices of the reference and target cameras, respectively. By combining \eqref{eq.Basis_Transformations} and \eqref{eq.camera_model}, we can reformulate the aforementioned three constraints into the following expressions:
\begin{equation}
    \begin{aligned}
        \text{X-Axis:}\ \ &{\boldsymbol{p}^{C}_1} = {\boldsymbol{p}^{C}_2} + s_1\boldsymbol{l}_1  \rightarrow \displaystyle\frac{y_1}{z_1} = \displaystyle\frac{y_2}{z_2} \\
    \rightarrow &\frac{v_1 - v_{o,1}}{f_{y,1}} = \frac{v_2 - v_{o,2}}{f_{y,2}},
    \label{eq.constraints_1}
    \end{aligned}
\end{equation}
\begin{equation}
    \begin{aligned}
        \text{Y-Axis:}\ \ &{\boldsymbol{p}^{C}_1} = {\boldsymbol{p}^{C}_2} + s_2\boldsymbol{l}_2  \rightarrow \displaystyle\frac{x_1}{z_1} = \displaystyle\frac{x_2}{z_2} \\
    \rightarrow &\frac{u_1 - u_{o,1}}{f_{x,1}} = \frac{u_2-u_{o,2}}{f_{x,2}},
    \label{eq.constraints_2}
    \end{aligned}
\end{equation}
\begin{equation}
    \begin{aligned}
        \text{Z-Axis:}\ \ &{\boldsymbol{p}^{C}_1} = {\boldsymbol{p}^{C}_2} + s_3\boldsymbol{l}_3  \rightarrow \displaystyle\frac{x_1}{y_1} = \displaystyle\frac{x_2}{y_2} \\
    \rightarrow &\frac{u_1 - u_{o,1}}{v_1-v_{o,1}} = \frac{u_2-u_{o,2}}{v_2-v_{o,2}},
    \label{eq.constraints_3}
    \end{aligned}
\end{equation}
which serve as the foundation for developing geometric metrics in the subsequent subsection.

\subsection{Geometric Metrics}
\label{sec.Geometric_Metrics}

Existing methods \cite{hansen2012online, zhao2024dive} formulate the energy function as the sum of absolute residuals calculated based on the constraint presented in \eqref{eq.constraints_1}. In this subsection, we comprehensively analyze the drawbacks of existing methods in terms of absolute residual accumulation, and introduce three geometric metrics defined to quantify the distances between the relative pose to be estimated and the basis transformations. 

\begin{figure*}[t]
\centering
\includegraphics[width=0.88\textwidth]{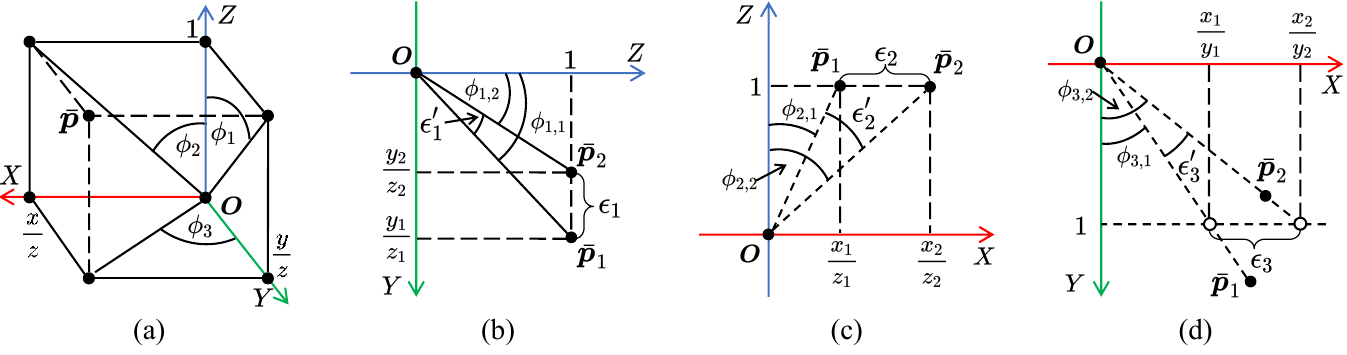}
\caption{
(a) Definition of $\boldsymbol{\Phi} = (\phi_1, \phi_2, \phi_3)$; (b) Comparison between ${\epsilon}_1$ and ${\epsilon}'_1$; (c) Comparison between ${\epsilon}_2$ and ${\epsilon}'_2$; (d) Comparison between ${\epsilon}_3$ and ${\epsilon}'_3$.
}
\label{fig.Euler}
\end{figure*}
\begin{figure}[t]
\centering
\includegraphics[width=0.99\linewidth]{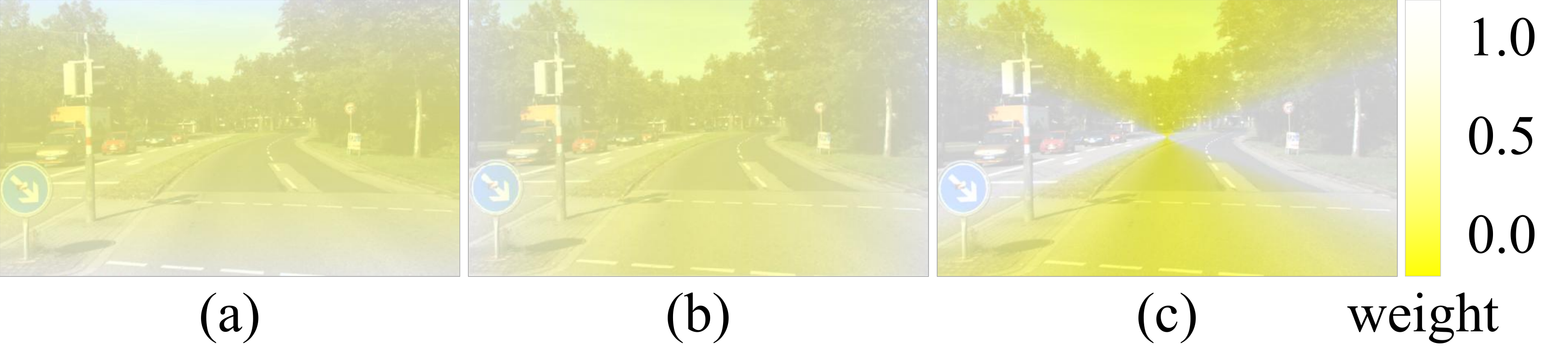}
\caption{Visualizations of the normalized implicit weights under different constraints: (a) Constraint given in \eqref{eq.constraints_1}; (b) Constraint given in \eqref{eq.constraints_2}; (c) Constraint given in \eqref{eq.constraints_3}.}
\label{fig.unfair}
\end{figure}

As illustrated in Fig.~\ref{fig.Euler}(a), $\bar{\boldsymbol{p}}_1$ and $\bar{\boldsymbol{p}}_2$, the normalized coordinates of a pair of correspondences with respect to depth, can be respectively expressed using $\boldsymbol{\Phi}_1 = (\phi_{1,1}, \phi_{2,1}, \phi_{3,1})^{\top}$ and $\boldsymbol{\Phi}_2 = (\phi_{1,2}, \phi_{2,2}, \phi_{3,2})^{\top}$ as follows:
\begin{equation}
\begin{cases}
    \bar{\boldsymbol{p}}_1 = \left(\displaystyle\frac{x_1}{z_1}, \displaystyle\frac{y_1}{z_1}, 1\right)^\top = (\tan(\phi_{2,1}), \tan(\phi_{1,1}), 1)^\top, \\[5pt]
    \bar{\boldsymbol{p}}_2 = \left(\displaystyle\frac{x_2}{z_2}, \displaystyle\frac{y_2}{z_2}, 1\right)^\top = (\tan(\phi_{2,2}), \tan(\phi_{1,2}), 1)^\top.
\end{cases}
    \label{eq.angles_1}
\end{equation}
These angles satisfy the following relationship:
\begin{equation}
    \frac{\tan(\phi_1)\tan(\phi_3)}{\tan(\phi_2)}=1.
    \label{eq.phiphiphi}
\end{equation}
Combining \eqref{eq.angles_1} and \eqref{eq.phiphiphi} results in the following expressions:
\begin{equation}
    \begin{cases}
    \displaystyle\frac{x_1}{y_1} = \frac{x_1}{z_1}\frac{z_1}{y_1} = \tan(\phi_{3,1}),\\[7pt]
    \displaystyle\frac{x_2}{y_2} = \frac{x_2}{z_2}\frac{z_2}{y_2} = \tan(\phi_{3,2}).
\end{cases}
    \label{eq.angles_2}
\end{equation}
According to \eqref{eq.angles_1} and \eqref{eq.angles_2}, an absolute residual vector $\boldsymbol{\epsilon}$, formulated based on the geometric constraints presented in \eqref{eq.constraints_1}-\eqref{eq.constraints_3}, can be expressed in terms of $\boldsymbol{\Phi}_1$ and $\boldsymbol{\Phi}_2$ as follows: 
\begin{equation}
\begin{aligned}
    \boldsymbol{\epsilon} &= \left(\left|\frac{y_1}{z_1} - \frac{y_2}{z_2}\right|, \left|\frac{x_1}{z_1} - \frac{x_2}{z_2}\right|, \left|\frac{x_1}{y_1} - \frac{x_2}{y_2}\right|\right) \\
    &= \left| \tan(\boldsymbol{\Phi}_1) - \tan(\boldsymbol{\Phi}_2) \right|.
\end{aligned}
        \label{eq.err_unfair}
\end{equation}
${\epsilon}_i$, the $i$-th element of $\boldsymbol{\epsilon}$, is illustrated in Fig. \ref{fig.Euler}(b)-(d), respectively. As the relative pose to be estimated nears alignment with the $i$-th basis transformation, $\Delta \phi_i = \lvert \phi_{i,1} - \phi_{i,2} \rvert$ approaches zero. The ratio of ${\epsilon}_i$ to $\Delta\phi_i$ can be formulated as follows:
\begin{equation}
\begin{aligned}
    \lim_{\Delta\phi_i \to 0}\frac{{\epsilon}_i}{\Delta\phi_i} &= \lim_{\phi_{i,2} \to \phi_{i,1}} \frac{\left| \tan(\phi_{i,1}) - \tan(\phi_{i,2}) \right|}{\Delta\phi_i} \\ &= \frac{1}{\cos^2(\phi_{i,1})},
\end{aligned}
\end{equation}
which leads to the following expression:
\begin{equation}
    {\epsilon}_i \approx \frac{\Delta\phi_i}{\cos^2(\phi_{i,1})},
\end{equation}
indicating that ${\epsilon}_i$ increases monotonically with respect to $\phi_{i,1}$. Therefore, the implicit weights of residuals currently in use are subject to pixel coordinates. To illustrate this bias, we visualize the normalized implicit weights under different constraints, as shown in Fig. \ref{fig.unfair}. Specifically, under the first constraint, the implicit weight increases along the vertical axis (see Fig. \ref{fig.unfair}(a)), peaking near the top and bottom image boundaries and reaching a minimum near the horizontal midline. Similarly, under the second constraint, the implicit weight increases along the horizontal axis (see Fig. \ref{fig.unfair}(b)), peaking near the left and right image boundaries and reaching a minimum near the vertical midline. Nonetheless, as illustrated in Fig. \ref{fig.unfair}(c), under the third constraint, the implicit weights along the horizontal midline reach the maximum, whereas those along the vertical midline reach the minimum. In addition, the implicit weights are symmetric about the image center and decrease monotonically as the line passing through the center rotates from the horizontal to the vertical midline. This phenomenon emphasizes the instability of previous relative pose estimation approaches, particularly when correspondences from specific regions, such as treetops or sky, are utilized.

To overcome this limitation, we adjust the three constraints in \eqref{eq.constraints_1}-\eqref{eq.constraints_3} using an inverse trigonometric function, as expressed as follows:
\begin{align}
\text{X-Axis:}&\ \ \arctan \frac{v_1 - v_{o,1}}{f_{y,1}} = \arctan \frac{v_2 - v_{o,2}}{f_{y,2}},\label{eq.new_constraint_1}\\
\text{Y-Axis:}&\ \ \arctan \frac{u_1 - u_{o,1}}{f_{x,1}} = \arctan \frac{u_2-u_{o,2}}{f_{x,2}},\label{eq.new_constraint_2}\\
\text{Z-Axis:}&\ \ \arctan \frac{u_1 - u_{o,1}}{v_1-v_{o,1}} = \arctan \frac{u_2-u_{o,2}}{v_2-v_{o,2}}\label{eq.new_constraint_3}.
\end{align}
The new absolute residual vector $\boldsymbol{\epsilon}'$ can therefore be expressed as follows:
\begin{equation}
    \boldsymbol{\epsilon}' = | \boldsymbol{\Phi}_1 - \boldsymbol{\Phi}_2 |=(\Delta\phi_1, \Delta\phi_2, \Delta\phi_3),
\end{equation}
which eliminates the influence of implicit weights in residual accumulation. Comparisons between the absolute residuals ${\epsilon}_i$ and ${\epsilon}_i'$ are also given in Fig. \ref{fig.Euler}(b)-(d).

In accordance with the new constraints defined in \eqref{eq.new_constraint_1}-\eqref{eq.new_constraint_3}, we introduce three geometric metrics that correspond to the three basis transformations as follows:
\begin{align}  
\text{X-Axis:} \ \ d_1 = \iint \left(\Delta \arctan \frac{v-v_o}{f}\right)^2 \, du \, dv,
\label{eq.distance_1}
\\
\text{Y-Axis:} \ \ d_2 = \iint \left(\Delta \arctan \frac{u-u_o}{f}\right)^2 \, du \, dv,
\label{eq.distance_2}
\\
\text{Z-Axis:} \ \ d_3 = \iint \left(\Delta \arctan \frac{u-u_o}{v-v_o}\right)^2 \, du \, dv.
\label{eq.distance_3}
\end{align}
The value of $d_i$ represents the distance between the actual relative pose and the $i$-th basis transformation. A smaller $d_i$ indicates a closer alignment between the actual relative pose and the $i$-th basis transformation. When exactly one of the three distances approaches zero, the actual relative pose aligns with the corresponding basis transformation. Additionally, it is worth noting that when the camera remains stationary, all three distances approach zero.

\subsection{The Birotation Solution}
\label{sec.The_Birotation_Solution}

Building on the above-mentioned basis transformations and geometric metrics, we introduce three birotation solutions to the relative pose estimation problem.

We use two rotation matrices, $\boldsymbol{R}_1$ and $\boldsymbol{R}_2$, to apply pure rotation transformations to the reference and target camera coordinate systems, respectively, in order to align them with the $i$-th basis transformation. This process can be formulated as follows:
\begin{equation}
    \begin{cases}
        \begin{aligned}
            {\boldsymbol{p}^{C}_1}{'}
            &=\boldsymbol{R}_1 \boldsymbol{p}^{C}_1 =z_1\boldsymbol{R}_1\bar{\boldsymbol{p}}_1,
        \end{aligned}
        \\
        \begin{aligned}
            {\boldsymbol{p}^{C}_2}{'}
            &=\boldsymbol{R}_2 \boldsymbol{p}^{C}_2
            =z_2\boldsymbol{R}_2 \bar{\boldsymbol{p}}_2,
        \end{aligned}
        \\
        \begin{aligned}
            {\boldsymbol{p}^{C}_1}{'}
            ={\boldsymbol{p}^{C}_2}{'} + s_i\boldsymbol{l}_i, \ \
            i \in \{1, 2, 3\},
        \end{aligned}
    \end{cases}
\label{eq.rectification}
\end{equation}
where $\boldsymbol{p}^{C'}_{1,2}$ denotes the transformed coordinates of $\boldsymbol{p}^C_{1,2}$ in the new camera coordinate systems. We refer to \eqref{eq.rectification} as the $i$-th birotation model for each $i \in \{1,2,3\}$, where the corresponding parameters $\boldsymbol{R}_1$, $\boldsymbol{R}_2$, $s_i$, and $\boldsymbol{l}_i$ together constitute the $i$-th birotation solution for the relative pose problem.

By substituting the $i$-th birotation model into the geometric metrics presented in \eqref{eq.distance_1}-\eqref{eq.distance_3}, the three geometric metrics can be reformulated as functions of $\boldsymbol{R}_1$ and $\boldsymbol{R}_2$, as follows:
\begin{equation}
     d_i = \iint \left(\arctan \frac{\boldsymbol{r}_{1,j}^{\top} \bar{\boldsymbol{p}}_1}{\boldsymbol{r}_{1,k}^{\top}\bar{\boldsymbol{p}}_1} - \arctan \frac{\boldsymbol{r}_{2,j}^{\top}\bar{\boldsymbol{p}}_2}{\boldsymbol{r}_{2,k}^{\top}\bar{\boldsymbol{p}}_2}\right)^2 \, du \, dv,
    \label{eq.di}
\end{equation}
where $(i\ j\ k)$ forms an even permutation \cite{dixon1996permutation} of the set $\{1, 2, 3\}$ when $i \in \{1, 3\}$, and an odd permutation when $i = 2$. Given the scaling factor $s_i$, the $i$-th birotation solution can be obtained by minimizing $d_i$.

\begin{figure*}[t]
\centering
\includegraphics[width=0.99\textwidth]{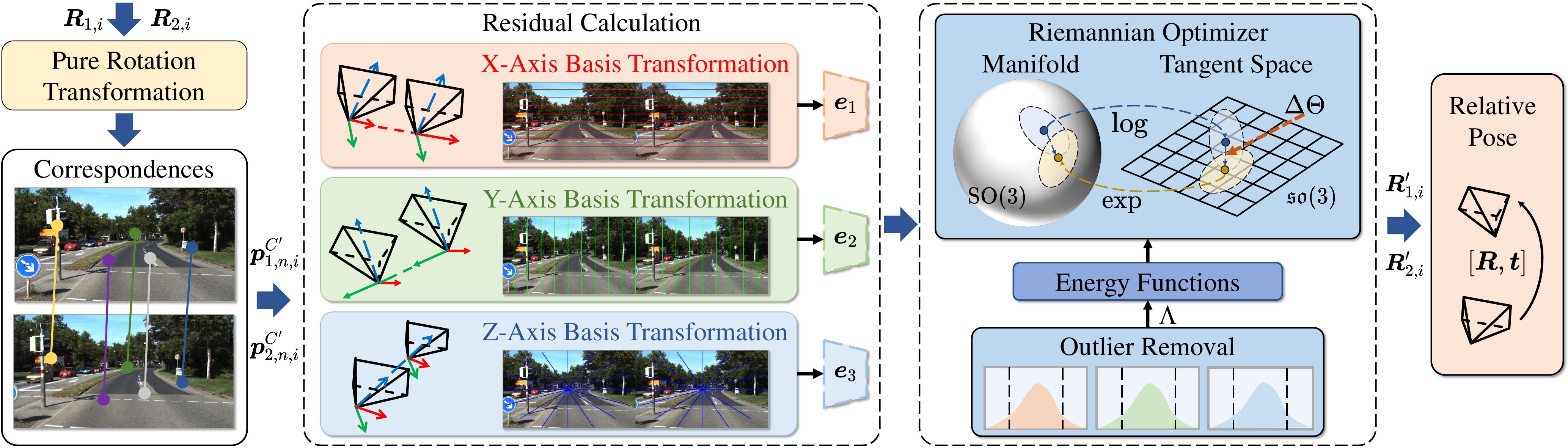}
\caption{
An overview of a single iteration in our proposed relative pose estimation algorithm. For each $i \in \{1,2,3\}$, two rotation matrices $\boldsymbol{R}_{1,i}$ and $\boldsymbol{R}_{2,i}$, obtained from the previous iteration, are used to rotate $N$ pairs of correspondences. The transformed correspondences $\boldsymbol{p}^{C'}_{1,n,i}$ and $\boldsymbol{p}^{C'}_{2,n,i}$ are then used to compute the residual vectors $\boldsymbol{e}_1$, $\boldsymbol{e}_2$, and $\boldsymbol{e}_3$ via \eqref{eq.e_in}. The residuals, along with the diagonal matrix $\boldsymbol{\Lambda}$ (used for outlier removal), are incorporated into the energy functions defined in \eqref{eq.E}. 
To optimize the energy, the rotation matrices $\boldsymbol{R}_{1, i}$ and $\boldsymbol{R}_{2, i}$ are mapped to the tangent space $\mathfrak{so}(3)$ via the logarithmic map, allowing the increment vector $\Delta\boldsymbol{\Theta}_i$ to be solved in closed form. The updated rotation matrices $\boldsymbol{R}_{1, i}$ and $\boldsymbol{R}_{2, i}$ are then obtained on the Riemannian manifold $\mathrm{SO(3)}$ through the exponential map. If convergence has not yet been achieved, $\boldsymbol{R}'_{1, i}$ and $\boldsymbol{R}'_{2, i}$ are passed to the next iteration; otherwise, they are used to derive the final relative pose via \eqref{eq.Rt}.
}
\label{fig.overview}
\end{figure*}
The relative pose $\boldsymbol{P}$ and the essential matrix $\boldsymbol{E}$ can be derived from the birotation solutions. By multiplying both sides of $\tilde{\boldsymbol{p}}^C_2 = \boldsymbol{P}\tilde{\boldsymbol{p}}^C_1$ with the matrix $\begin{pmatrix}
    \boldsymbol{R}_2 & \boldsymbol{0}
\end{pmatrix}$, we have the following expression:
\begin{equation}
    \boldsymbol{R}_2\boldsymbol{p}^C_2 = \boldsymbol{R}_2\boldsymbol{R}\boldsymbol{p}^C_1 + \boldsymbol{R}_2\boldsymbol{t}.
    \label{eq.RP}
\end{equation}
According to the $i$-th birotation model defined in \eqref{eq.rectification}, the following relationship holds:
\begin{equation}
    \boldsymbol{p}^{C'}_2 = \boldsymbol{R}_2\boldsymbol{p}^C_2 = \boldsymbol{p}^{C'}_1 - s_i\boldsymbol{l}_i = 
    \boldsymbol{R}_1\boldsymbol{p}^C_1 - s_i\boldsymbol{l}_i.
    \label{eq.RRT}
\end{equation}
By combining \eqref{eq.RP} and \eqref{eq.RRT}, the rotation matrix $\boldsymbol{R}$ and translation vector $\boldsymbol{t}$ can be determined from the $i$-th birotation solution as follows:
\begin{equation}
    \begin{cases}
        \boldsymbol{R} = \boldsymbol{R}^{\top}_2\boldsymbol{R}_1,\\
        \boldsymbol{t} = -s_i\boldsymbol{R}^{\top}_2\boldsymbol{l}_i = -s_i\boldsymbol{r}_{2,i}.
    \end{cases}
\label{eq.Rt}
\end{equation}
Ideally, the three birotation solutions yield an identical relative pose. The essential matrix $\boldsymbol{E}$ can then be derived from the birotation solutions as follows:
\begin{equation}
    \begin{aligned}
        \boldsymbol{E} &= \frac{1}{\| t \|_2}[\boldsymbol{t}]_{\times}\boldsymbol{R} = -[\boldsymbol{r}_{2,i}]_{\times} \boldsymbol{R}^{\top}_2\boldsymbol{R}_1 \\
        &=\begin{cases}
            (\boldsymbol{0}, \boldsymbol{r}_{2,3}, -\boldsymbol{r}_{2,2})\boldsymbol{R}_1, & \text{if} \ i = 1, \\       
            (-\boldsymbol{r}_{2,3}, \boldsymbol{0}, \boldsymbol{r}_{2,1})\boldsymbol{R}_1, & \text{if} \ i = 2, \\
            (\boldsymbol{r}_{2,2}, -\boldsymbol{r}_{2,1}, \boldsymbol{0})\boldsymbol{R}_1, & \text{if} \ i = 3. 
            \end{cases} \\
        & = \begin{cases}
            \boldsymbol{r}_{2,3}\boldsymbol{r}^{\top}_{1,2} - \boldsymbol{r}_{2,2}\boldsymbol{r}^{\top}_{1,3}, & \text{if} \ i = 1, \\ 
            \boldsymbol{r}_{2,1}\boldsymbol{r}^{\top}_{1,3} - \boldsymbol{r}_{2,3}\boldsymbol{r}^{\top}_{1,1}, & \text{if} \ i = 2, \\ 
            \boldsymbol{r}_{2,2}\boldsymbol{r}^{\top}_{1,1} - \boldsymbol{r}_{2,1}\boldsymbol{r}^{\top}_{1,2}, & \text{if} \ i = 3. \\ 
        \end{cases}
    \end{aligned}
\end{equation}

However, it should be noted here that all three birotation solutions are not unique, as each solution has six DoFs (three from each rotation matrix), while the relative pose problem with ambiguous scale has only five DoFs. Specifically, rotating both the reference and target camera coordinate systems by the same angle around the translation vector does not change their relative pose, which geometrically demonstrates the redundant DoF in each birotation solution. Thus, additional geometric constraints should be incorporated into our proposed relative pose estimation algorithm, detailed in the next subsection, to resolve this ambiguity.

\subsection{Relative Pose Estimation Algorithm}
\label{sec.Relative_Pose_Estimation_Algorithm}
According to the constraints presented in \eqref{eq.new_constraint_1}-\eqref{eq.new_constraint_3} and the birotation models defined in \eqref{eq.rectification}, we can obtain three $N$-entry residual vectors using $N$ pairs of correspondences, as expressed as follows:
\begin{equation}    
    \boldsymbol{e}_i(\boldsymbol{\theta}_{1,i}, \boldsymbol{\theta}_{2,i}) = (e_{i,1},e_{i,2},\dots,e_{i,N})^\top, \ \ i \in \{1,2,3\},
    \label{eq.errors}
\end{equation}
where 
\begin{equation}
    \begin{aligned}
        e_{i,n} =   
        \arctan \frac{\boldsymbol{r}_{1,j,i}^{\top} \bar{\boldsymbol{p}}_{1,n}}{\boldsymbol{r}_{1,k,i}^{\top}\bar{\boldsymbol{p}}_{1,n}} - \arctan \frac{\boldsymbol{r}_{2,j,i}^{\top}\bar{\boldsymbol{p}}_{2,n}}{\boldsymbol{r}_{2,k,i}^{\top}\bar{\boldsymbol{p}}_{2,n}}, \\ n \in [1,N]\cap\mathbb{Z},
    \end{aligned}
    \label{eq.e_in}
\end{equation}
$\boldsymbol{\theta}_{1,i}$ and $\boldsymbol{\theta}_{2,i}$ denote the rotation vectors corresponding to the rotation matrices $\boldsymbol{R}_{1,i}$ and $\boldsymbol{R}_{2,i}$ in the $i$-th birotation model, respectively, $\bar{\boldsymbol{p}}_{1,n}$ and $\bar{\boldsymbol{p}}_{2,n}$ represent the normalized coordinates of the $n$-th pair of correspondences with respect to depth, $\boldsymbol{r}_{1,j,i}$ and $\boldsymbol{r}_{1,k,i}$ denote the $j$-th and $k$-th row vector of $\boldsymbol{R}_{1,i}$, respectively, and $\boldsymbol{r}_{2,j,i}$ and $\boldsymbol{r}_{2,k,i}$ are the $j$-th and $k$-th row vector of $\boldsymbol{R}_{2,i}$, respectively. $\hat{d}_i$, the discretized approximation of $d_i$ defined in \eqref{eq.di}, can be yielded by accumulating residuals defined in \eqref{eq.e_in} as follows:
\begin{equation}
    \hat{d}_i = \boldsymbol{e}^{\top}_i(\boldsymbol{\theta}_{1,i}, \boldsymbol{\theta}_{2,i})\boldsymbol{\Lambda} \boldsymbol{e}_i(\boldsymbol{\theta}_{1,i}, \boldsymbol{\theta}_{2,i}), \ \ i \in \{1,2,3\},
    \label{eq.discrete}
\end{equation}
where $\boldsymbol{\Lambda}$ is a diagonal matrix in which each element $\lambda_n$ adaptively takes a value of either 0 or 1 to indicate the reliability of the $n$-th pair of correspondences. By incorporating a regularization term into $\hat{d}_i$, the complete energy function for the $i$-th birotation model can be formulated as follows:
\begin{equation}
        E_i(\boldsymbol{\theta}_{1,i}, \boldsymbol{\theta}_{2,i}) = \boldsymbol{e}^{\top}_i \boldsymbol{\Lambda} \boldsymbol{e}_i + \alpha R(\boldsymbol{\theta}_{1,i}, \boldsymbol{\theta}_{2,i}), \ \ i\in\{1,2,3\},
    \label{eq.E}
\end{equation}
where $R(\boldsymbol{\theta}_{1,i}, \boldsymbol{\theta}_{2,i}) = \| \boldsymbol{\theta}_{1,i} \|^2_2 + \| \boldsymbol{\theta}_{2,i} \|^2_2$ denotes the $L_2$ regularization term, and $\alpha > 0$ represents the regularization parameter. The regularization term is incorporated to reduce the energy function's non-convexity and ensure the solution's uniqueness. 

The $i$-th birotation solution can be yielded by minimizing (\ref{eq.E}) using our newly designed iterative optimization approach. Specifically, based on \eqref{eq.Rt}, the three birotation solutions are initialized using the prior rotation $\boldsymbol{R}_{init}$ and translation $\boldsymbol{t}_{init}$ as follows:
\begin{align}
    &\begin{cases}
        \boldsymbol{r}_{2,1,1} = -\frac{\boldsymbol{t}_{init}}{\|\boldsymbol{t}_{init}\|}, \\
        \boldsymbol{r}_{2,2,1} = \boldsymbol{l}_3 \times \boldsymbol{r}_{2,1,1}, \\
        \boldsymbol{r}_{2,3,1} = \boldsymbol{r}_{2,1,1} \times \boldsymbol{r}_{2,2,1}, \\
    \end{cases} \\
    &\begin{cases}
        \boldsymbol{r}_{2,2,2} = -\frac{\boldsymbol{t}_{init}}{\|\boldsymbol{t}_{init}\|}, \\
        \boldsymbol{r}_{2,3,2} = \boldsymbol{l}_1 \times \boldsymbol{r}_{2,2,2}, \\
        \boldsymbol{r}_{2,1,2} = \boldsymbol{r}_{2,2,2} \times \boldsymbol{r}_{2,3,2},
    \end{cases}\\
    &\begin{cases}
        \boldsymbol{r}_{2,3,3} = -\frac{\boldsymbol{t}_{init}}{\|\boldsymbol{t}_{init}\|}, \\
        \boldsymbol{r}_{2,1,3} = \boldsymbol{l}_2 \times \boldsymbol{r}_{2,3,3}, \\
        \boldsymbol{r}_{2,2,3} = \boldsymbol{r}_{2,3,3} \times \boldsymbol{r}_{2,1,3},
    \end{cases}\\
    &\boldsymbol{R}_{2,i} = (\boldsymbol{r}_{2,1,i}, \boldsymbol{r}_{2,2,i}, \boldsymbol{r}_{2,3,i})^\top, \\
    &\boldsymbol{R}_{1,i} = \boldsymbol{R}_{2,i}\boldsymbol{R}_{init}, \ \ i \in \{1,2,3\}.
\end{align}

An overview of a single iteration of the optimization process is illustrated in Fig.~\ref{fig.overview}. We first derive a first-order approximation of $\boldsymbol{e}_i$, which is then applied to the energy function $E_i$ as follows:
\begin{equation}
   E_i \approx (\boldsymbol{e}_i + \boldsymbol{J}^{\top}_i\Delta\boldsymbol{\Theta}_i)^{\top}\boldsymbol{\Lambda}(\boldsymbol{e}_i + \boldsymbol{J}^{\top}_i\Delta\boldsymbol{\Theta}_i) + \alpha R,
   \label{eq.E'}
\end{equation}
where $\Delta\boldsymbol{\Theta}_i = (\Delta\boldsymbol{\theta}^{\top}_{1,i}, \Delta\boldsymbol{\theta}^{\top}_{2,i})^{\top}$ denotes the increment vector at each iteration, and 
\begin{equation}
    \boldsymbol{J}_i = \begin{pmatrix}
        \displaystyle\frac{\partial e_{i,1}}{\partial  \boldsymbol{\theta}_{1,i}}, \displaystyle\frac{\partial e_{i,2}}{\partial  \boldsymbol{\theta}_{1,i}}, \cdots, \displaystyle\frac{\partial e_{i,N}}{\partial  \boldsymbol{\theta}_{1,i}} \\
        \displaystyle\frac{\partial e_{i,1}}{\partial  \boldsymbol{\theta}_{2,i}}, \displaystyle\frac{\partial e_{i,2}}{\partial  \boldsymbol{\theta}_{2,i}}, \cdots, \displaystyle\frac{\partial e_{i,N}}{\partial  \boldsymbol{\theta}_{2,i}}
    \end{pmatrix}^\top,
\end{equation}
is a matrix storing the gradients of $\boldsymbol{e}_i$ with respect to $\boldsymbol{\theta}_{1,i}$ and $\boldsymbol{\theta}_{2,i}$ computed using the left perturbation model \cite{ling2016high, zhao2024dive}, which avoids the need to compute the Baker-Campbell-Hausdorff Jacobian matrix at each iteration. Under the definitions of $\boldsymbol{p}^{C'}_{1,n,i} = \boldsymbol{R}_{1,i} \boldsymbol{p}^C_{1,n} = (x'_{1,n,i},y'_{1,n,i},z'_{1,n,i})^\top$ and $\boldsymbol{p}^{C'}_{2,n,i} = \boldsymbol{R}_{2,i} \boldsymbol{p}^C_{2,n} = (x'_{2,n,i}, y'_{2,n,i}, z'_{2,n,i})^\top$, we can derive the following expressions:
\begin{align}
    \begin{cases}
        \frac{\partial e_{1,n}}{\partial \boldsymbol{\theta}_{1,1}} &= 
        \begin{pmatrix}
            -1 ,
            \displaystyle\frac{x'_{1,n,1} y'_{1,n,1}}{{y'_{1,n,1}}^2 + {z'_{1,n,1}}^2}, 
            \displaystyle\frac{x'_{1,n,1} z'_{1,n,1}}{{y'_{1,n,1}}^2 + {z'_{1,n,1}}^2}
        \end{pmatrix}, \\[12pt]
        \frac{\partial e_{1,n}}{\partial \boldsymbol{\theta}_{2,1}} &=
        \begin{pmatrix}
            1 ,
            -\displaystyle\frac{x'_{2,n,1} y'_{2,n,1}}{{y'_{2,n,1}}^2 + {z'_{2,n,1}}^2},
            -\displaystyle\frac{x'_{2,n,1} z'_{2,n,1}}{{y'_{2,n,1}}^2 + {z'_{2,n,1}}^2}
        \end{pmatrix},
    \end{cases}\\
    \begin{cases}
            \frac{\partial e_{2,n}}{\partial \boldsymbol{\theta}_{1,2}} &= \begin{pmatrix}
        -\displaystyle\frac{x'_{1,n,2} y'_{1,n,2}}{{x'_{1,n,2}}^2 + {z'_{1,n,2}}^2},
        1,
        -\displaystyle\frac{y'_{1,n,2} z'_{1,n,2}}{{x'_{1,n,2}}^2 + {z'_{1,n,2}}^2}
        \end{pmatrix}, \\[12pt]
            \frac{\partial e_{2,n}}{\partial \boldsymbol{\theta}_{2,2}} &= 
            \begin{pmatrix}
            \displaystyle\frac{x'_{2,n,2} y'_{2,n,2}}{{x'_{2,n,2}}^2 + {z'_{2,n,2}}^2},
        -1 ,
        \displaystyle\frac{y'_{2,n,2} z'_{2,n,2}}{{x'_{2,n,2}}^2 + {z'_{2,n,2}}^2}
        \end{pmatrix},
    \end{cases}\\
    \begin{cases}
    \frac{\partial e_{3,n}}{\partial \boldsymbol{\theta}_{1,3}} &=
        \begin{pmatrix}
            \displaystyle\frac{x'_{1,n,3} z'_{1,n,3}}{{x'_{1,n,3}}^2 + {y'_{1,n,3}}^2}, 
            \displaystyle\frac{y'_{1,n,3} z'_{1,n,3}}{{x'_{1,n,3}}^2 + {y'_{1,n,3}}^2},
            -1
        \end{pmatrix}, \\[12pt]
        \frac{\partial e_{3,n}}{\partial \boldsymbol{\theta}_{2,3}}
        &= \begin{pmatrix}
        -\displaystyle\frac{x'_{2,n,3} z'_{2,n,3}}{{x'_{2,n,3}}^2 + {y'_{2,n,3}}^2},    
        -\displaystyle\frac{y'_{2,n,3} z'_{2,n,3}}{{x'_{2,n,3}}^2 + {y'_{2,n,3}}^2},
        1
        \end{pmatrix}.
        \end{cases}
\end{align}
Setting the derivative of \eqref{eq.E'} with respect to $\Delta\boldsymbol{\Theta}_i$ to zero yields the following expression: 
\begin{equation}
 \Delta\boldsymbol{\Theta}_i = -({\boldsymbol{J}_i}\boldsymbol{\Lambda}{\boldsymbol{J}_i}^{\top} + \alpha \boldsymbol{I})^{-1}{\boldsymbol{J}_i}\boldsymbol{\Lambda}\boldsymbol{e}_i.
\label{eq.dtheta}
\end{equation} 
Subsequently, $\boldsymbol{R}_{1,i}$ and $\boldsymbol{R}_{2,i}$ are updated using the left perturbation model, as follows: 
\begin{equation}
    \begin{cases}
        \boldsymbol{R}'_{1,i} = \exp([\Delta\boldsymbol{\theta}_{1,i}]_{\times})\boldsymbol{R}_{1,i},\\
        \boldsymbol{R}'_{2,i} = \exp([\Delta\boldsymbol{\theta}_{2,i}]_{\times})\boldsymbol{R}_{2,i},
    \end{cases}
    \label{eq.update}
\end{equation}
where $\boldsymbol{R}'_{1,i}$ and $\boldsymbol{R}'_{2,i}$ denote the updated rotation matrices after one iteration. The iterative process terminates when either the value or the change rate of $\frac{\hat{d}_i}{N}$ falls below predefined thresholds. After the iteration terminates, the final solution is selected from the three birotation solutions based on the following criteria:
\begin{equation}
    \boldsymbol{R}_1, \boldsymbol{R}_2 = \boldsymbol{R}_{1,\hat{i}}, \boldsymbol{R}_{2,\hat{i}}, \quad \hat{i} = \arg\min_i \beta_i \hat{d}_i,
    \label{eq.select}
\end{equation}
where weighting coefficients $(\beta_1, \beta_2, \beta_3)$ can be empirically adjusted to prioritize one of the three birotation solutions, which can greatly enhance the reliability of the final solution. The relative rotation $\boldsymbol{R}$ and the line along which the translation vector $\boldsymbol{t}$ lies can then be derived from $\boldsymbol{R}_1$, $\boldsymbol{R}_2$, and their corresponding index $\hat{i}$ using \eqref{eq.Rt}.   

In addition, although the absolute value of the scaling factor $s_{\hat{i}}$ is unknown, its sign can be determined based on the transformed coordinates $\boldsymbol{p}^{C'}_{1,n,\hat{i}}$ and $\boldsymbol{p}^{C'}_{2,n,\hat{i}}$ using the following expression when $z'_{1,n,\hat{i}}$ and $z'_{2,n,\hat{i}}$ are positive:
\begin{equation}
    \begin{cases}
       s_{\hat{i}} > 0, \quad \text{if} \quad \hat{i} = 1 \quad \text{and} \quad \frac{x'_{1,n,\hat{i}}}{z'_{1,n,\hat{i}}} > \frac{x'_{2,n,\hat{i}}}{z'_{2,n,\hat{i}}}, \\[8pt]
       s_{\hat{i}} > 0, \quad \text{if} \quad \hat{i} = 2 \quad \text{and} \quad \frac{y'_{1,n,\hat{i}}}{z'_{1,n,\hat{i}}} > \frac{y'_{2,n,\hat{i}}}{z'_{2,n,\hat{i}}}, \\[8pt]
       s_{\hat{i}} > 0, \quad \text{if} \quad \hat{i} = 3 \quad \text{and} \quad \left| \frac{x'_{1,n,\hat{i}}}{z'_{1,n,\hat{i}}} \right| < \left| \frac{x'_{2,n,\hat{i}}}{z'_{2,n,\hat{i}}} \right|, \\[8pt]
       s_{\hat{i}} < 0, \quad \text{otherwise}.
    \end{cases}
\end{equation}
When the depth of either $\boldsymbol{p}^{C'}_{1,n,\hat{i}}$ or $\boldsymbol{p}^{C'}_{2,n,\hat{i}}$ is negative, the direction of the translation is determined by computing the depths of both $\boldsymbol{p}^C_{1,n}$ and $\boldsymbol{p}^C_{2,n}$. The translation yielding two positive depths is selected as the valid solution. With convergence-related parameters fixed, the computational complexity of our relative pose estimation algorithm remains $\mathcal{O}(N)$.

\section{Experiments}
\label{sec.Experiments}

Sect.~\ref{sec.Implementation_Details} provides implementation details on the proposed algorithm and the methods used for comparison. Sect.~\ref{sec.Datasets} introduces the four public datasets utilized for performance comparison, along with the datasets newly created for ablation studies and robustness analyses. The evaluation metrics are detailed in Sect.~\ref{sec.Evaluation_Metrics}. Sect.~\ref{sec.Comparisons_with_SoTA_Algorithms} presents extensive quantitative and qualitative comparisons between the proposed method and previous SoTA approaches. In Sect.~\ref{sec.Ablation_Study}, ablation studies are conducted to evaluate the contributions of the energy functions and the effectiveness of the constraint adjustments. Finally, Sect.~\ref{sec.Virtual_Experiments} analyzes the robustness of our approach under different initialization strategies, correspondence noise levels, and matching error rates.

\subsection{Implementation Details}
\label{sec.Implementation_Details}

\subsubsection{Correspondence Matching}
For both comparisons with SoTA methods and ablation studies, we adopt LoFTR \cite{sun2021loftr}, trained on the MegaDepth dataset \cite{li2018megadepth}, to obtain dense correspondences. This is primarily because $d_i$ in \eqref{eq.di} can be closely approximated with $\hat{d}_i$ in \eqref{eq.discrete} when correspondences are dense. Additionally, we employ SIFT \cite{lowe2004distinctive} in conjunction with LightGlue \cite{lindenberger2023lightglue}, also trained on the MegaDepth dataset, to further evaluate the robustness of our algorithm over other relative pose estimation algorithms under sparse correspondence conditions.

\begin{table*}[t!]
\settablefont
\centering
\caption{Quantitative comparisons of relative pose algorithms on the ScanNet \cite{dai2017scannet} dataset. }
\begin{tabular*}{0.9\textwidth}{@{\extracolsep\fill}r c c c c c c c c}
    \toprule
    \multirow{2}{*}{Algorithm} & \multicolumn{4}{c}{Dense Correspondence} & \multicolumn{4}{c}{Sparse Correspondence}\\
    \cmidrule{2-5} \cmidrule{6-9}
     & @$1^{\circ}$ & @$3^{\circ}$ & @$5^{\circ}$ & @$10^{\circ}$ & @$1^{\circ}$ & @$3^{\circ}$ & @$5^{\circ}$ & @$10^{\circ}$\\
    \midrule
    {Nist{\'e}r} \cite{nister2004efficient} &0.56&6.84&15.24&31.49 & \underline{0.74} & 7.82&16.54&33.44 \\
    Helmke {{\etal}} \cite{helmke2007essential} & 0.05&0.89&2.76&7.42& 0.08&2.21&5.51&14.11  \\
    Zhao \cite{zhao2020efficient} & 0.00&0.04&0.23&1.49 & 0.00&0.07&0.40&2.12\\
    Ling {\etal} \cite{ling2016high} & \underline{0.83} & \underline{8.39} & \underline{16.55} & \underline{32.15} & 0.58& \underline{8.11} & \underline{17.18} & \textbf{34.56}  \\
    Zhao {\etal} \cite{zhao2024dive} & 0.63&6.34&12.83&26.16& 0.74 & 6.62 & 13.93 & 29.48 \\
    \textbf{Ours} & \textbf{1.08} & \textbf{9.56} & \textbf{18.65} & \textbf{33.08} & \textbf{1.05} & \textbf{8.53} & \textbf{17.75} & \underline{33.94} \\
    \hline
\end{tabular*}
\label{tab.ScanNet}
\end{table*}

\begin{table*}[t!]
\settablefont
\centering
\caption{Quantitative comparisons of relative pose algorithms on the YFCC100M \cite{thomee2016yfcc100m} dataset.} 
\begin{tabular*}{0.9\textwidth}{@{\extracolsep\fill}r c c c c c c c c}
    \toprule
    \multirow{2}{*}{Algorithm} 
     & \multicolumn{4}{c}{Dense Correspondence} & \multicolumn{4}{c}{Sparse Correspondence}\\
    \cmidrule{2-5} \cmidrule{6-9}
     & @$1^{\circ}$ & @$3^{\circ}$ & @$5^{\circ}$ & @$10^{\circ}$ & @$1^{\circ}$ & @$3^{\circ}$ & @$5^{\circ}$ & @$10^{\circ}$\\
    \midrule
    Nist{\'e}r \cite{nister2004efficient} &6.57&26.84&41.65&62.01& 6.95&26.74&41.18&\underline{60.74} \\
    Helmke {\etal} \cite{helmke2007essential} &0.84&6.03&11.58&22.80& 1.50&10.17&18.68&33.75 \\
    Zhao \cite{zhao2020efficient} &0.04&0.76&2.44&9.04 & 0.07&0.96& 2.86& 9.78 \\
    Ling {\etal} \cite{ling2016high} & \underline{9.50} & \underline{31.81} & \textbf{46.59} & \textbf{65.51} & \textbf{10.00} & \textbf{31.13} & \textbf{45.13} & \textbf{63.53} \\
    Zhao {\etal} \cite{zhao2024dive} & 5.04&21.25&33.36&52.27& 5.61&21.85&33.71&52.19 \\
    \textbf{Ours} &\textbf{10.15} & \textbf{32.02} & \underline{45.89} & \underline{63.45} & \underline{9.49} & \underline{29.19} & \underline{41.67} & 58.22 \\
    \hline
\end{tabular*}
\label{tab.YFCC100M}
\end{table*}

\begin{table*}[h]
\settablefont
\centering
\caption{Quantitative comparisons of relative pose algorithms on the pure rotation dataset. }
\begin{tabular*}{\textwidth}{@{\extracolsep\fill}crcccccccc}
    \toprule
    \multirow{2}{*}{Rotation Angle} & \multirow{2}{*}{Algorithm} &  \multicolumn{4}{c}{Dense Correspondence} &  \multicolumn{4}{c}{Sparse Correspondence}\\
    \cmidrule{3-6} \cmidrule{7-10}
    & & @$1^{\circ}$ & @$3^{\circ}$ & @$5^{\circ}$ & @$10^{\circ}$ & @$1^{\circ}$ & @$3^{\circ}$ & @$5^{\circ}$ & @$10^{\circ}$ \\
    \midrule 
    \multirow{7}{*}{$\leq 5^{\circ}$} 
    & Nist{\'e}r \cite{nister2004efficient} & 47.90&52.06&52.89&53.52& 64.87&70.69&71.86&72.73 \\
    & Helmke {\etal} \cite{helmke2007essential} &47.92&52.59&53.57&54.44& 67.24&73.91&75.47&76.77 \\
    & Zhao \cite{zhao2020efficient} & 89.00&96.33&97.80&98.90&90.13&96.71&98.03&99.01 \\
    & Ling {\etal} \cite{ling2016high} & 88.75&96.25&97.75&98.87&88.82&96.27&97.76&98.88  \\
    & Zhao {\etal} \cite{zhao2024dive} & 68.36&74.09&75.24&76.10&76.11&82.19&83.41&84.32  \\
    & Hartley \cite{hartley2003multiple} & \underline{91.45} & \underline{97.15} & \underline{98.29} & \underline{99.14} & \textbf{92.83}&\textbf{97.61}&\textbf{98.57}&\textbf{99.28} \\
    & \textbf{Ours}  & \textbf{91.77} & \textbf{97.26} & \textbf{98.35} & \textbf{99.18} & \underline{90.54} & \underline{96.85} & \underline{98.11} & \underline{99.05} \\
    \midrule
    \multirow{7}{*}{$\leq 10^{\circ}$} 
    & Nist{\'e}r \cite{nister2004efficient} &53.47&58.09&59.02&59.71 & 66.30&72.51&73.75&74.68 \\
    & Helmke {\etal} \cite{helmke2007essential} & 52.02&56.99&58.00&58.75  & 68.76&75.76&77.37&78.69\\
    & Zhao \cite{zhao2020efficient} & 89.55&96.52&\textbf{97.91}&\textbf{98.96}  & 89.69&96.42&97.76&98.77\\
    & Ling {\etal} \cite{ling2016high} & 88.87 & 96.29 & 97.77 & \underline{98.89} & 88.94 & 96.31 & 97.79 & \underline{98.89} \\
    & Zhao {\etal} \cite{zhao2024dive} & 69.56&75.69&76.92&77.83  & 77.34&83.81&85.10&86.06 \\
    & Hartley \cite{hartley2003multiple} & \textbf{90.80} & \textbf{96.63} & \underline{97.80} & 98.68 & \textbf{91.64} & \textbf{96.91} & \underline{97.97} & 98.76 \\
    & \textbf{Ours}  & \underline{90.67} & \underline{96.59} & 97.78 & 98.66 & \underline{90.63} & \underline{96.88} & \textbf{98.13} & \textbf{99.06} \\
    \midrule
    \multirow{7}{*}{$\leq 15^{\circ}$} 
    & Nist{\'e}r \cite{nister2004efficient} &54.00&58.72&59.66&60.37& 68.67&75.69&77.09&78.14 \\
    & Helmke {\etal} \cite{helmke2007essential} &51.75&57.20&58.39&59.43  &72.44&80.18&81.83&83.08\\
    & Zhao \cite{zhao2020efficient} & 87.54&94.50&95.90&96.94& 82.82&89.20&90.48&91.44 \\
    & Ling {\etal} \cite{ling2016high} & 88.83&\textbf{96.28} &\textbf{97.77} &\textbf{98.88}&88.90&\textbf{96.30}&\textbf{97.78}&\textbf{98.89}\\
    & Zhao {\etal} \cite{zhao2024dive} & 69.97&76.72&78.07&79.08  & 80.28&87.01&88.36&89.37 \\
    & Hartley \cite{hartley2003multiple} & \underline{89.72} &95.53&96.69&97.56  & \textbf{90.24} & \underline{96.15} & 97.33 & 98.22\\
    & \textbf{Ours}  & \textbf{90.03} & \underline{96.23} & \underline{97.47} & \underline{98.40} & \underline{89.75} & 96.14 & \underline{97.41} & \underline{98.37} \\
    \hline
\end{tabular*}
\label{tab.pure_rotation}
\end{table*}

\subsubsection{Compared SoTA Algorithms}

We first compare our algorithm with the classical five-point algorithm used in conjunction with RANSAC \cite{nister2004efficient} (implemented in OpenCV with default parameters), which is an explicit solver whose output is then used as the initialization for subsequent local optimization methods. These local optimization methods include the manifold-based method \cite{helmke2007essential}, the method that directly estimates rotation and translation \cite{ling2016high}, and the birotation-based method \cite{zhao2024dive}. Moreover, we also use the published code of the QCQP-based global optimization method \cite{zhao2020efficient} to conduct comparative experiments. In addition, the homography estimation method \cite{hartley2003multiple}, as implemented in OpenCV (all parameters are set by default), is also used for further comparison under the pure rotation condition. In the monocular visual odometry task, we also compare our method with two SoTA data-driven methods, SC-Depth \cite{bian2021unsupervised} and SCIPaD \cite{feng2024scipad}, both trained on sequences 00–08 of the KITTI Odometry dataset \cite{geiger2013vision}.

\subsubsection{Parameter Settings in Our Algorithm}
The regularization parameter $\alpha$ defined in \eqref{eq.E} is set to $1 \times 10^{-3}$. The coefficients $(\beta_1, \beta_2, \beta_3)$ in \eqref{eq.select} are empirically set to $(0.25, 1, 1)$ in the stereo camera extrinsic calibration experiment, $(1, 1, 0.25)$ in the monocular visual odometry experiment, and the default values $(1, 1, 1)$ in all other experiments. This is because the ground-truth relative poses in the stereo camera extrinsic calibration and monocular visual odometry tasks typically align more closely with the first and third basis transformations defined in \eqref{eq.Basis_Transformations}, respectively. Our method is also initialized using the output of the five-point algorithm with RANSAC \cite{nister2004efficient}. In addition, the upper quartile method \cite{huber1992robust} is employed in each iteration to compute $\boldsymbol{\Lambda}$ in \eqref{eq.E} and remove outliers. The iteration terminates when either the value of $\frac{\hat{d}_i}{N}$ falls below $10^{-8}$ or its rate of change falls below $10^{-6}$.

\begin{table*}[t]
\centering
\caption{Quantitative comparisons of relative pose estimation algorithms for stereo camera extrinsic calibration. Given that the rotation error is generally small, its unit is reported in $10^{-3}$ radian in the table.}
\settablefont
    \begin{tabular}{@{\extracolsep\fill}C{1cm}cR{2cm}C{0.8cm}C{0.8cm}C{0.8cm}C{0.8cm}C{0.8cm}C{0.8cm}C{0.8cm}C{0.8cm}C{0.8cm}C{0.8cm}}
		\toprule
           {Scenes} & {Correspondence} & {Algorithm} & $\Delta\bar{\theta}_1 $ & $\Delta\bar{\theta}_2 $ & $\Delta\bar{\theta}_3 $ & $\Delta\bar{t}_1 $ & $\Delta\bar{t}_2 $ & $\Delta\bar{t}_3 $  & @1$^{\circ}$ & @3$^{\circ}$ & @5$^{\circ}$ & @10$^{\circ}$ \\ 
    \hline
    \multirow{12}{*}{Indoor} & \multirow{6}{*}{Dense}
    & Nist{\'e}r \cite{nister2004efficient} & 1.85& 6.07& 1.77 & 6.02& 6.52& 11.23 & 5.47& 26.30	& 42.40	& 64.50 \\
    && Helmke {\etal} \cite{helmke2007essential} &8.61& 8.77& 7.01 & 17.33& 12.30& 29.34 & 0.37 & 6.35	& 13.99	& 34.02\\
    && Zhao \cite{zhao2020efficient}  & 2.67& 13.74& 3.13 & 61.06& 15.74& 75.79 &0.00	& 4.99	& 16.18	& 29.89 \\
    && Ling {\etal} \cite{ling2016high}  & \underline{0.37} & \underline{2.51} & \underline{0.49} & \underline{5.77} & 2.44& 8.85 & 30.21	& 60.10	& 72.95	& 84.06\\
    && Zhao {\etal} \cite{zhao2024dive} & 41.53& 4.11& 6.28 & 6.69& \underline{2.22} & \underline{3.79} & \underline{32.08} & \underline{66.01} & \underline{77.34} & \underline{86.74}\\
    && \textbf{Ours} & \textbf{0.27} & \textbf{2.14} & \textbf{0.43} & \textbf{0.70} & \textbf{1.52} & \textbf{2.85} & \textbf{39.10} & \textbf{70.74}	& \textbf{81.41}	& \textbf{90.25} \\
    \cline{2-13}
    &\multirow{6}{*}{Sparse}
    & Nist{\'e}r \cite{nister2004efficient} & 2.01& 7.52& 2.24 & 11.98& 6.29& 13.60 &6.44	& 25.48	& 40.79	& 62.34 \\
    && Helmke {\etal} \cite{helmke2007essential} & 2.65& 7.93& 3.82 & 15.89& 13.08& 24.95& 0.63	& 7.63	& 16.28	& 34.78\\
    && Zhao \cite{zhao2020efficient}  & 2.74& 13.34& 3.42 & 51.61& 15.65& 67.92 & 0.00	& 5.28	& 17.18	& 31.65\\
    && Ling {\etal} \cite{ling2016high}  & \underline{0.40} & \underline{2.43} & \underline{0.43} & \underline{7.73} & \underline{2.54} & 10.97& 29.00	& 57.31	& 69.27	& 81.31\\
    && Zhao {\etal} \cite{zhao2024dive} & 41.09& 6.59& 4.97 & 13.47& 2.79& \underline{5.20} & \underline{35.55}	& \underline{68.01}	& \underline{78.28}	& \underline{86.85} \\
    && \textbf{Ours} & \textbf{0.25} & \textbf{2.17} & \textbf{0.38} & \textbf{2.85} & \textbf{1.66} & \textbf{4.19} & \textbf{36.67} & \textbf{70.41}	& \textbf{80.60}	& \textbf{89.47} \\
    \hline
    \multirow{12}{*}{Outdoor}&\multirow{6}{*}{Dense}
    & Nist{\'e}r \cite{nister2004efficient} & 1.31& 6.40& 1.03 & 2.36& 3.35& 15.82 & 2.86 & 16.19	& 28.36	& 52.39 \\
    && Helmke {\etal} \cite{helmke2007essential} & 2.24& 6.18& 3.19 & 6.51& 7.50& 17.51 &0.77 & 8.28	& 19.38	& 44.64 \\
    && Zhao \cite{zhao2020efficient} & 1.10& 1.24& 0.66 & 4.48& 6.00& 11.83 & 0.00	& 11.72	& 40.89	& 67.99	\\
    && Ling {\etal} \cite{ling2016high}  & 0.40& 1.40& 0.29 & \underline{0.40} & 1.55 & 7.46 &13.55	& 39.69	& 56.50	& 76.31 \\
    && Zhao {\etal} \cite{zhao2024dive} & \underline{0.26} & \textbf{0.87} & \underline{0.22} & 1.32 & \underline{0.87} & \underline{1.67} & \underline{44.71} & \underline{80.12} & \underline{88.07} & \underline{94.04} \\
    && \textbf{Ours} & \textbf{0.23} & \underline{0.96} & \textbf{0.20} & \textbf{0.12} & \textbf{0.64} & \textbf{1.30} & \textbf{52.15} & \textbf{83.69} & \textbf{90.22} & \textbf{95.11} \\
    \cline{2-13}
    &\multirow{6}{*}{Sparse}
    & Nist{\'e}r \cite{nister2004efficient} & 1.22& 5.91& 1.15 & 2.37& 3.43& 14.89 & 4.71 & 19.61	& 32.86	& 55.13 \\
    && Helmke {\etal} \cite{helmke2007essential} & 2.54& 6.83& 3.76 & 8.55& 6.79& 17.42 &1.84	& 10.31	& 22.49	& 46.63 \\
    && Zhao \cite{zhao2020efficient} & 0.43& 1.14& 0.31 & 1.36& 6.01& 9.34 &0.19	& 11.86	& 40.22	& 68.71 \\
    && Ling {\etal} \cite{ling2016high}  & 0.39& 1.28& 0.34 & \underline{0.46} & 1.62& 7.58 &16.98	& 42.38	& 57.84	& 76.46 \\
    && Zhao {\etal} \cite{zhao2024dive} & \textbf{0.25} & \underline{0.83} & \underline{0.24} & 1.31 & \underline{0.83} & \underline{1.41} & \underline{52.36}	& \underline{83.12}	& \underline{89.87}	& \underline{94.93} \\
    && \textbf{Ours} & \textbf{0.25} & \textbf{0.77} & \textbf{0.23} & \textbf{0.10} & \textbf{0.55} & \textbf{0.91} & \textbf{64.36} & \textbf{88.06} & \textbf{92.83}	& \textbf{96.42} \\
    \hline
    \end{tabular}
    \label{tab.calib}
  \end{table*}

\subsection{Datasets}
\label{sec.Datasets}

\subsubsection{Public Datasets}
\label{sec.Public_Datasets}

\begin{itemize}
    \item The \textbf{ScanNet} dataset \cite{dai2017scannet} is a large-scale dataset, consisting of 2.5 million indoor RGB-D image pairs with ground-truth camera poses.  Following the study \cite{sarlin2020superglue}, we select 1,500 image pairs to evaluate the performance of relative pose estimation algorithms in indoor scenarios.
    
    \item The \textbf{YFCC100M} dataset \cite{thomee2016yfcc100m} is a large-scale public multimedia collection containing approximately 99.2 million images with ground-truth camera poses. Following the study in \cite{sarlin2020superglue}, we select 4,000 image pairs to evaluate the performance of relative pose estimation algorithms in outdoor scenarios.
    
    \item The \textbf{KITTI Odometry} dataset \cite{geiger2013vision} consists of well-rectified stereo video sequences and ground-truth trajectories collected from 22 real-world driving scenarios. Existing data-driven approaches, including the two networks we compare \cite{bian2021unsupervised,feng2024scipad}, are typically trained on sequences 0 to 8 and evaluated on sequences 9 and 10. In our experiments, we evaluate all explicitly programming-based algorithms on sequences 0 to 10. Although this may introduce a slight bias when compared to data-driven approaches, the experimental results presented in Sect. \ref{sec.Practical_Applications} demonstrate the superiority of our method over them. Additionally, we select 300 consecutive video frames from KITTI Odometry sequence 0, where the relative pose closely aligns with the third basis transformation, to conduct an ablation study. The results demonstrate the advantage of using three energy functions for relative pose estimation over a single energy function. The images used in the ablation study are publicly available for result reproduction.
    
    \item The \textbf{KITTI Raw} dataset \cite{geiger2013vision} consists of raw video sequences and ground-truth trajectories collected from 56 real-world driving scenarios. In our experiments, we select around 500 images as reference frames and generate three subsets of target frames by applying random rotations with angles of up to $5^{\circ}$, $10^{\circ}$, and $15^{\circ}$, respectively. These subsets are used to evaluate the performance of relative pose estimation algorithms when only rotation is involved. The images used for the pure rotation experiment are publicly available for result reproduction. 
\end{itemize}

\begin{figure*}[t]
    \centering
    \includegraphics[width=0.99\textwidth]{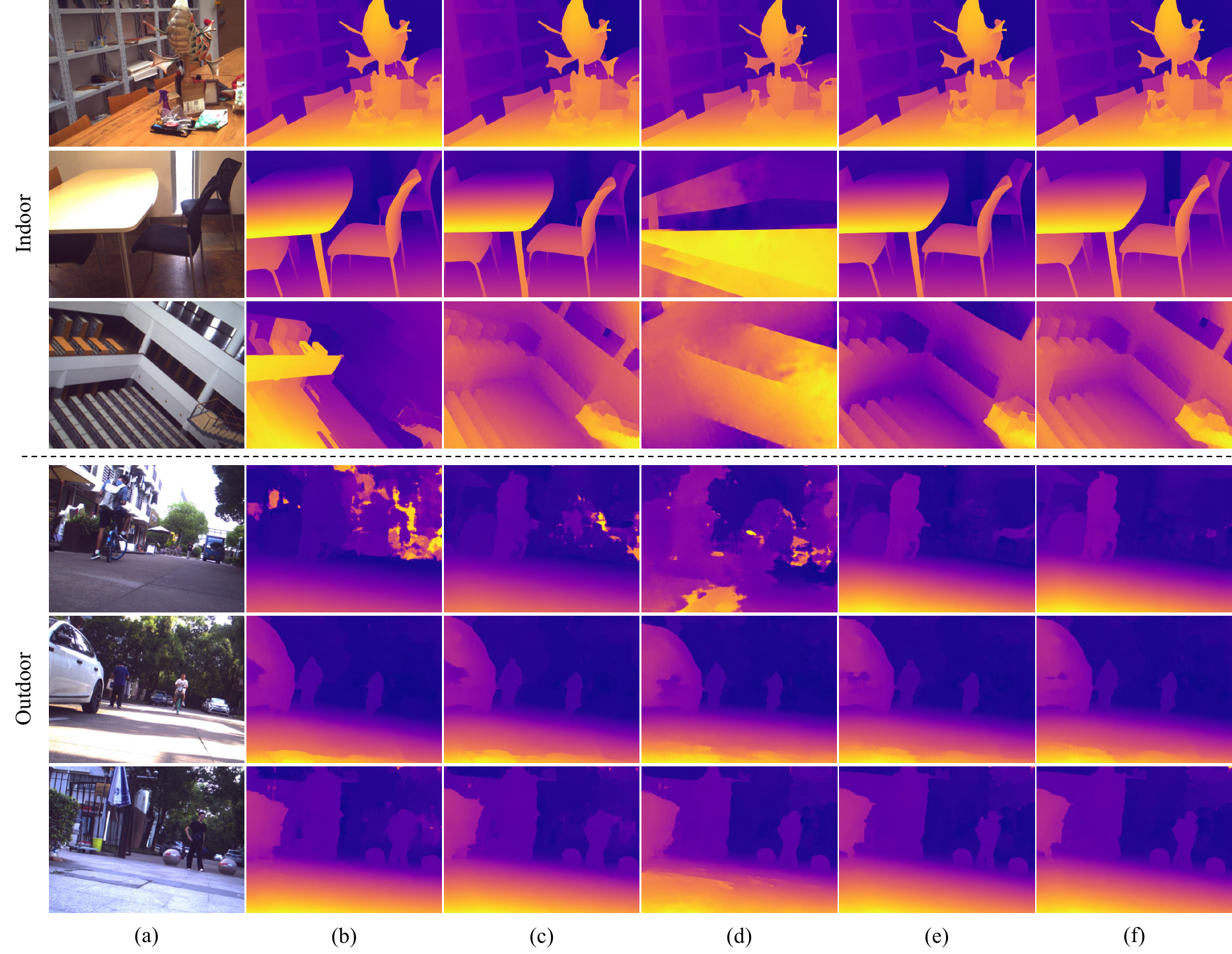}
    \caption{Qualitative comparisons of relative pose algorithms on the stereo camera extrinsic calibration task: (a) Left images rectified based on the ground-truth extrinsic parameters; (b-f) Disparity maps generated from stereo images rectified based on the extrinsic parameters estimated using \cite{nister2004efficient}, \cite{ling2016high}, \cite{zhao2020efficient}, \cite{zhao2024dive}, and our algorithm, respectively.}
    \label{fig.disp}
\end{figure*}

\begin{figure*}[t]
    \centering
    \includegraphics[width=0.99\textwidth]{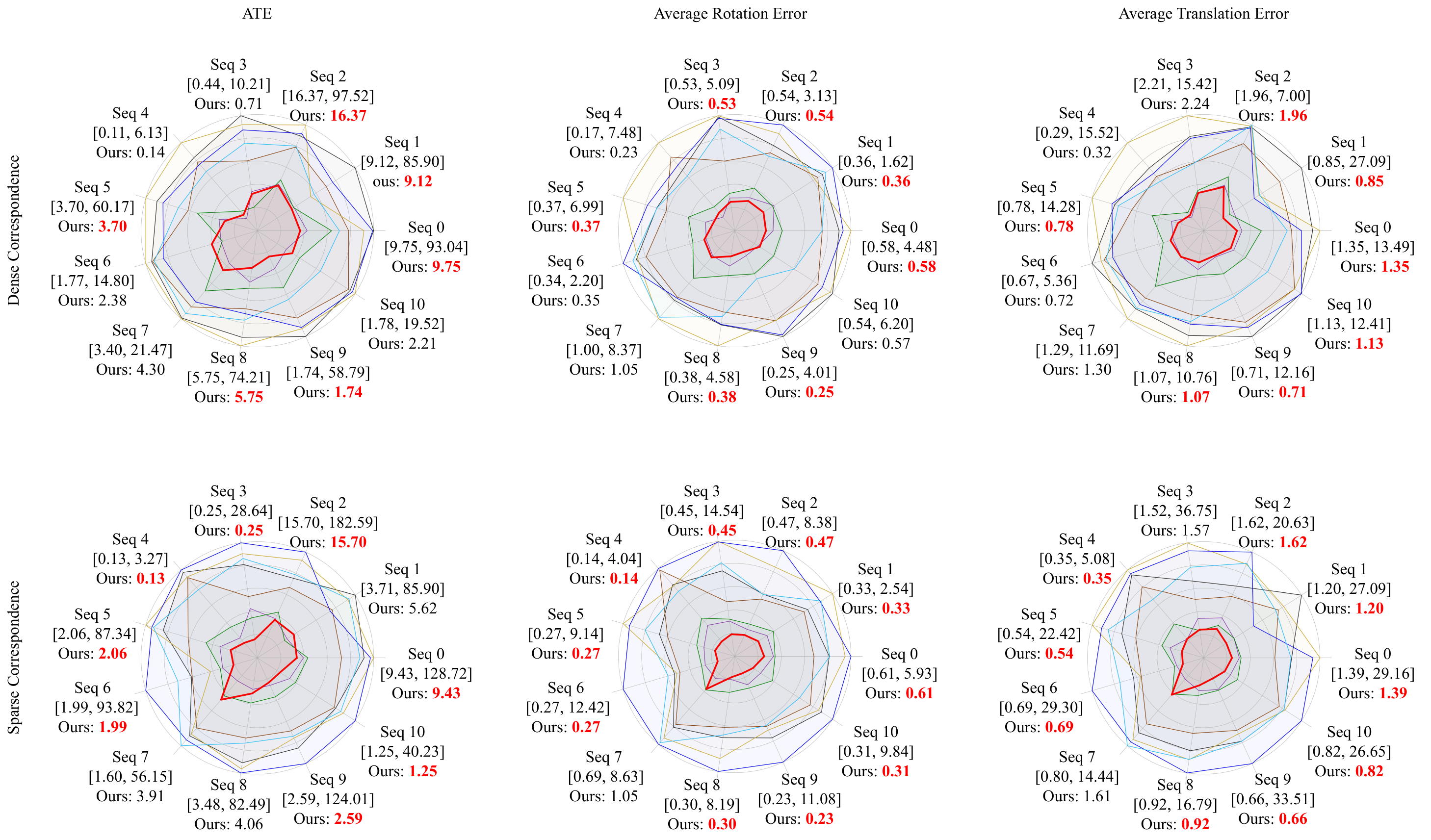}
   \caption{Quantitative comparisons on the KITTI Odometry dataset, where ``Seq $i$'' refers to the $i$-th sequence. The figures present the $[\min,\max]$ ranges along with our results, which are highlighted in red when optimal. A smaller area enclosed by lines indicates better performance of the corresponding algorithm. \protect\tikz[baseline=-0.5ex]{\protect\draw[SC-Depth, thick] (0,0) -- (0.35,0);} SC-Depth, \protect\tikz[baseline=-0.5ex]{\protect\draw[SCIPaD, thick] (0,0) -- (0.35,0);} SCIPaD, \protect\tikz[baseline=-0.5ex]{\protect\draw[Nist, thick] (0,0) -- (0.35,0);} Nist{\'e}r \cite{nister2004efficient}, \protect\tikz[baseline=-0.5ex]{\protect\draw[Helmke, thick] (0,0) -- (0.35,0);} Helmke {\etal} \cite{helmke2007essential}, \protect\tikz[baseline=-0.5ex]{\protect\draw[zhao2024, thick] (0,0) -- (0.35,0);} Zhao {\etal} \cite{zhao2024dive}, \protect\tikz[baseline=-0.5ex]{\protect\draw[zhao2020, thick] (0,0) -- (0.35,0);} Zhao \cite{zhao2020efficient}, \protect\tikz[baseline=-0.5ex]{\protect\draw[ling, thick] (0,0) -- (0.35,0);} Ling {\etal} \cite{ling2016high}, and \protect\tikz[baseline=-0.5ex]{\protect\draw[red, very thick] (0,0) -- (0.35,0);} ours.}
    \label{fig.odometry_lidar}
\end{figure*}

\begin{figure*}[t]
    \centering
    \includegraphics[width=0.99\textwidth]{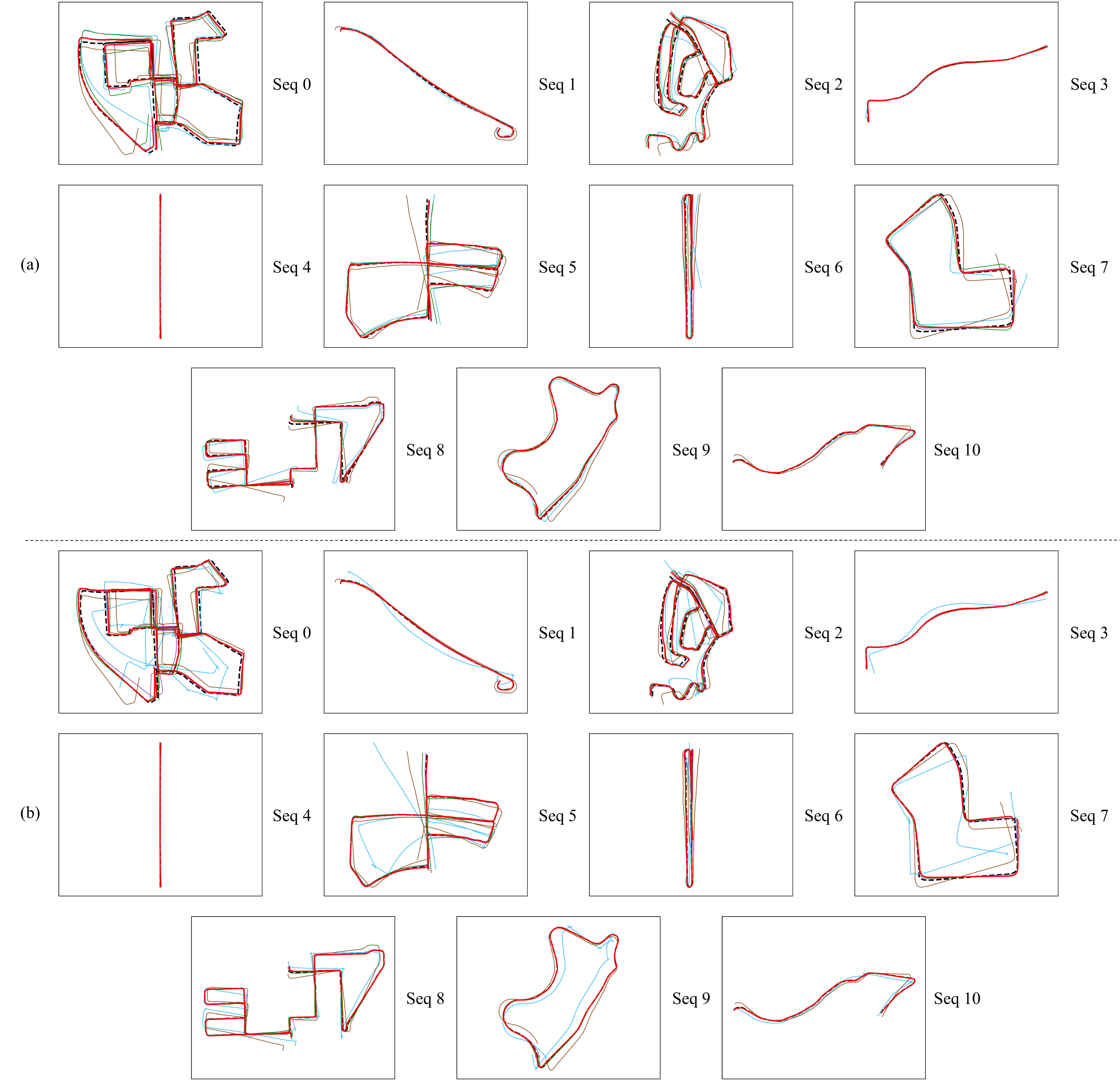}
    \caption{Comparisons of estimated trajectories on the KITTI Odometry dataset:
    (a) Estimated trajectories when correspondences are dense; (b) Estimated trajectories when correspondences are sparse. \protect\tikz[baseline=-0.5ex]{\protect\draw[dashed, very thick] (0,0) -- (0.35,0);} Ground truth, \protect\tikz[baseline=-0.5ex]{\protect\draw[SCIPaD, thick] (0,0) -- (0.35,0);} SCIPaD, \protect\tikz[baseline=-0.5ex]{\protect\draw[Nist, thick] (0,0) -- (0.35,0);} Nist{\'e}r \cite{nister2004efficient}, \protect\tikz[baseline=-0.5ex]{\protect\draw[zhao2020, thick] (0,0) -- (0.35,0);} Zhao \cite{zhao2020efficient}, \protect\tikz[baseline=-0.5ex]{\protect\draw[ling, thick] (0,0) -- (0.35,0);} Ling {\etal} \cite{ling2016high}, and \protect\tikz[baseline=-0.5ex]{\protect\draw[red, very thick] (0,0) -- (0.35,0);} ours.}
    \label{fig.odometry}
\end{figure*}

\subsubsection{New Real-World Datasets}
\label{sec.Created_Real_World_Datasets}
As discussed above, ablation studies are required to compare the performance of our algorithm when using three energy functions versus a single energy function. Therefore, in addition to the 300 images selected from the KITTI Odometry dataset, we also create two additional datasets (each containing 300 image pairs), in which the cameras are positioned on the left and right, and top and bottom, respectively. Given that a relatively ideal experimental setup may not comprehensively reflect the performance of our algorithm, we introduce slight perturbations to the stereo rig in practice. Moreover, we use the same stereo rig to collect an additional 600 image pairs (300 indoor and 300 outdoor) for the stereo camera extrinsic online calibration experiment.

\subsubsection{New Synthetic Datasets}
\label{sec.Created_Synthetic_Datasets}

We also create two synthetic datasets to further evaluate the robustness of our algorithm. We first generate a synthetic 3-D point cloud by uniformly sampling points within a cubic volume in the world coordinate system. All cameras have the same intrinsic parameters. The reference camera coordinate system is aligned with the world coordinate system, whereas the target camera coordinate systems are randomly generated. The synthetic 3-D points are then projected onto each camera’s image plane to generate corresponding 2-D pixel coordinates. The details of the two synthetic datasets are as follows: 

\begin{itemize}
    \item The first synthetic dataset is created by adding Gaussian noise to the 2-D correspondences, with the standard deviation $\sigma$ ranging from 0 to 2 pixels in increments of 0.02 pixel. For each noise level, we generate 100 pairs of reference and target views, each observing 200 3-D points.

    \item The second synthetic dataset is created under varying correspondence matching error rates, ranging from 0 to 0.3 in increments of 0.01. Correct correspondences are perturbed with slight Gaussian noise ($\sigma = 0.1$ pixel), while erroneous correspondences are perturbed with stronger Gaussian noise ($\sigma = 10$ pixels) to simulate outliers. For each error rate, we generate 100 pairs of reference and target views, each observing 200 3-D points.
\end{itemize}
All the synthetic datasets are publicly available for result reproduction. 

\subsection{Evaluation Metrics}
\label{sec.Evaluation_Metrics}
To quantify the algorithm's performance with respect to individual motion components, we follow the studies \cite{ling2016high} and \cite{zhao2024dive} to compute the average absolute rotation error $\Delta\bar{\boldsymbol{\theta}} = \frac{1}{M}\sum^{M}_{m = 1}\lvert \hat{\boldsymbol{\theta}}_m - \boldsymbol{\theta}^{\ast} \rvert = (\Delta\bar{\theta}_1, \Delta\bar{\theta}_2, \Delta\bar{\theta}_3)$ in radians and average absolute translation error $\Delta\bar{\boldsymbol{t}} = \frac{1}{M}\sum^{M}_{m=1}\lvert \hat{\boldsymbol{t}}_m - \boldsymbol{t}^{\ast} \rvert = (\Delta\bar{t}_1, \Delta\bar{t}_2, \Delta\bar{t}_3)$ in millimeters, respectively, where $M$ denotes the total number of frames used for evaluation, $\hat{\boldsymbol{\theta}}_m$ denotes the $m$-th rotation vector derived from the estimated rotation matrix $\hat{\boldsymbol{R}}_m$, $\boldsymbol{\theta}^{\ast}$ represents the ground-truth rotation vector, and $\hat{\boldsymbol{t}}_m$ and $\boldsymbol{t}^{\ast}$ denote the $m$-th estimated and ground-truth translation vectors, respectively. To further quantify the overall performance of the algorithm in rotation and translation estimation, we follow the study \cite{zhao2020efficient} to compute the rotation error $\varepsilon_{\boldsymbol{r}}= \arccos\frac{1}{2}(\text{tr}(\boldsymbol{R}^{*\top} \hat{\boldsymbol{R}}) - 1)$ and translation error $\varepsilon_{\boldsymbol{t}}= \arccos\left(\left|\frac{\boldsymbol{t}^{*}}{\|\boldsymbol{t}^{*}\|} \cdot \frac{\hat{\boldsymbol{t}}}{\|\hat{\boldsymbol{t}}\|}\right|\right)$, both measured in degrees. Moreover, we follow the study \cite{sarlin2020superglue} to quantify the algorithm's overall relative pose estimation performance using the area under the cumulative pose error curve (AUC) in percents, computed at thresholds of $1^\circ$, $3^\circ$, $5^\circ$, and $10^\circ$, respectively, where the pose error is defined as the maximum of $\varepsilon_{\boldsymbol{r}}$ and $\varepsilon_{\boldsymbol{t}}$. In this article, $@\psi$ denotes the AUC value at the threshold $\psi$, with higher values indicating better performance. It is important to note that when only rotational motion is involved, the AUC is calculated solely based on the rotation error. Additionally, when evaluating the algorithm's performance on the monocular visual odometry task, we adopt commonly used metrics as presented in \cite{geiger2013vision}, including the average translational error $e_{\boldsymbol{t}}$ (\%), the rotational error $e_{\boldsymbol{r}}$ (${}^{\circ}/100$m), and the absolute trajectory error (ATE) in meters \cite{sturm2012benchmark}. To ensure fair comparisons, following the study \cite{feng2024scipad}, the scales of all estimated trajectories are aligned with the ground truth using 7-DoF optimization.

\subsection{Comparisons with SoTA Algorithms}
\label{sec.Comparisons_with_SoTA_Algorithms}

This subsection first evaluates the performance of our algorithm under challenging conditions. Additional experiments on the pure rotation dataset are then conducted to validate the effectiveness of our method when only rotations are involved. Finally, experiments on stereo camera extrinsic online calibration and monocular visual odometry are conducted to demonstrate the practicality of our algorithm in specific tasks.

\subsubsection{Performance under Challenging Conditions}
\label{sec.Performance_Evaluation_under_Challenging_Relative_Pose_Conditions}

Since this study aims to extend the existing birotation-based approach \cite{zhao2024dive} to general relative pose problems, it is necessary to evaluate the performance of our method on two public datasets, ScanNet and YFCC100M, where the ground-truth relative poses can be significantly different from any of the three predefined basis transformations, making correspondence matching considerably more challenging. Following the study \cite{sarlin2020superglue}, we compute the AUC (\%) across varying thresholds $\psi$ ($^\circ$) to quantify the performance of relative pose estimation algorithms on both datasets. 

As shown in Table~\ref{tab.ScanNet}, our algorithm achieves SoTA performance on the ScanNet dataset for both dense and sparse correspondences, which demonstrates the effectiveness of our method in indoor scenarios. Moreover, the outdoor relative pose estimation results presented in Table~\ref{tab.YFCC100M} indicate that our approach performs on par with the best-performing method \cite{ling2016high} when correspondences are dense, and achieves the second-best performance when correspondences are sparse. Notably, while all methods use the output of the algorithm \cite{nister2004efficient} as the initial relative pose, the previous birotation-based approach \cite{zhao2024dive} performs significantly worse than the algorithm \cite{nister2004efficient}, whereas our method achieves substantial improvements. This underscores the limitation of the previous study \cite{zhao2024dive}, which was primarily designed for relatively ideal stereo vision models, and demonstrates the effectiveness of our approach in broadening the applicability of the birotation solution by introducing two additional basis transformations. In addition, although our method performs slightly worse than the algorithm presented in \cite{ling2016high} in outdoor scenarios, this discrepancy may be attributed to the outlier removal strategy incorporated in their approach. Unlike the upper quartile method used in our algorithm, their strategy depends on an empirically tuned hyperparameter (extensive hyperparameter tuning experiments are conducted in this study to ensure fair comparisons), which limits its practicality across diverse datasets.

\subsubsection{Performance under Pure Rotation Conditions}
\label{sec.Pure_rotation_estimation}

Pure rotation represents a corner case in relative pose estimation. Nonetheless, most existing algorithms developed for general relative pose problems rarely take this corner case into consideration. Therefore, it is essential to evaluate the performance of existing methods under pure rotation conditions using the curated pure rotation dataset detailed in Sect.~\ref{sec.Public_Datasets}. The results are shown in Table~\ref{tab.pure_rotation}.

As expected, the approaches \cite{nister2004efficient, helmke2007essential} that estimate the essential matrix perform poorly across all thresholds, which validates the limitations of these methods in this corner case, as discussed in Sect.~\ref{sec.intro}. While the approach \cite{zhao2020efficient} also estimates the essential matrix, it achieves significantly better performance, primarily because it incorporates statistical analyses to explicitly identify pure rotation cases. In addition, the algorithm \cite{ling2016high}, which directly estimates relative poses, demonstrates strong robustness in pure rotation scenarios. We attribute this to the normalization of the translation vector and the decoupled optimization of rotation and translation in their method.

Moreover, the previous birotation-based method \cite{zhao2024dive} struggles to stably solve pure rotation problems, primarily due to its sensitivity to the initialization of relative poses induced by the penalty term. In contrast, our proposed algorithm demonstrates superior robustness across all evaluation thresholds, regardless of whether the correspondences are dense or sparse. In addition, our approach performs on par with the homography estimation approach \cite{hartley2003multiple}, which is specifically designed for pure rotation scenarios, further validating the accuracy of our algorithm in this corner case.

\subsubsection{Performance on Two Specific Tasks}
\label{sec.Practical_Applications}

Additionally, we conduct extensive experiments to demonstrate the applicability of our relative pose estimation algorithm to two specific computer vision tasks: stereo camera extrinsic online calibration and monocular visual odometry.

In addition to AUC, we follow the studies \cite{ling2016high} and \cite{zhao2024dive} to compute the average absolute rotation and translation errors in the stereo camera extrinsic online calibration experiment. As shown in Table~\ref{tab.calib}, our algorithm consistently achieves SoTA performance in both indoor and outdoor scenarios. Notably, it increases the three translation components by approximately $73\%$, $33\%$, and $22\%$ in indoor scenarios, and by $74\%$, $30\%$, and $28\%$ in outdoor scenarios. As expected, the translation vectors estimated by the algorithm \cite{ling2016high} are occasionally unsatisfactory, primarily due to the significant non-convexity of its energy function shown in \eqref{eq.Rt_solution}. Moreover, although the previous birotation-based method \cite{zhao2024dive}, specifically designed for stereo camera extrinsic online calibration, achieves relatively high AUC, it yields unexpectedly large average absolute rotation and translation errors in indoor scenarios due to outliers in the estimated relative poses. These outliers are primarily caused by the inherent non-uniqueness of the birotation solution, as discussed in Sect.~\ref{sec.discussion}. In contrast, our method ensures a unique solution by incorporating a regularization term into the energy function defined in \eqref{eq.E}. Furthermore, in outdoor scenarios, our method outperforms the method \cite{zhao2024dive}, primarily because the adjustments in constraints defined in \eqref{eq.new_constraint_1}-\eqref{eq.new_constraint_3} eliminate the influence of implicit weights in residual accumulation, as described in Sect. \ref{sec.Geometric_Metrics}. In addition, we also provide qualitative comparisons of different relative pose estimation approaches by visualizing disparity maps estimated using pre-trained deep stereo networks \cite{chang2018pyramid, lipson2021raft}. These disparity maps are generated from stereo images rectified using the extrinsic parameters estimated by different relative pose estimation algorithms. As illustrated in Fig.~\ref{fig.disp}, our method produces disparity maps with fewer erroneous regions compared to other approaches, which further validates its greater accuracy in both outdoor and indoor scenarios.

Quantitative comparisons of relative pose estimation algorithms for monocular visual odometry are depicted in Fig.~\ref{fig.odometry_lidar}, where our method achieves SoTA performance across all three matrices. Specifically, when correspondences are dense, our method achieves improvements of around $3\%$ in ATE, $7\%$ in $e_{\boldsymbol{r}}$, and $6\%$ in $e_{\boldsymbol{t}}$, while these gains increase to around $6\%$, $21\%$, and $17\%$ when using sparse correspondences. Notably, our algorithm outperforms existing approaches on Seqs 0, 2, 5, 8, and 9, which were collected along complex motion trajectories. These results underscore the robustness of our method for monocular visual odometry, particularly in challenging scenarios. As expected, the previous birotation-based method \cite{zhao2024dive} performs poorly on this task, as ground-truth relative poses are mostly aligned with the third basis transformation, making the approach originally designed for relatively ideal stereo vision model infeasible. Moreover, the two SoTA data-driven approaches underperform several explicit programming-based methods, as current learning-based visual odometry frameworks lack sufficient geometric constraints during training to ensure accurate relative pose estimation. We also visualize the trajectories obtained by the five top-performing approaches to further compare their monocular visual odometry performance. As illustrated in Fig. \ref{fig.odometry}, our approach yields trajectories that closely align with the ground truth in all sequences when using both dense and sparse correspondences, which demonstrates its high accuracy and robustness in the monocular visual odometry task.

\subsection{Ablation Studies}
\label{sec.Ablation_Study}

\begin{table}[t!]
\settablefont
\centering
\caption{Ablation study on energy function contributions.}
\begin{tabular}{cccccccc}
   \toprule
   \shortstack{Ground-Truth\\Relative Pose}
    & $E_1$ & $E_2$ & $E_3$ & @$1^{\circ}$ & @$3^{\circ}$ & @$5^{\circ}$ & @$10^{\circ}$\\    
    \midrule
    \multirow{4}{*}{Left-right} 
    & $\checkmark$ & & & \textbf{55.94} & \textbf{85.02 } & \textbf{91.01} & \textbf{95.51} \\
    & & $\checkmark$ & &  54.35 & 83.37 & 89.62 & 94.54 \\
    & & & $\checkmark$ & 0.19 &  0.29&  0.30 &  0.54\\
    & $\checkmark$ & $\checkmark$ & $\checkmark$ & \textbf{55.94} & \textbf{85.02 } & \textbf{91.01} & \textbf{95.51} \\
    \midrule
    \multirow{4}{*}{Top-bottom} 
    & $\checkmark$ & & & 3.69	 & 20.66 & 32.16 & 48.49 \\ 
    & & $\checkmark$ & & 8.84 & \textbf{44.67} & \textbf{63.69} & \textbf{81.36} \\
    & & & $\checkmark$ & 1.10	  & 5.33	 &  7.85 & 11.79 \\
     & $\checkmark$ & $\checkmark$ & $\checkmark$ & \textbf{8.93} & 44.54 & 63.61 & 80.94 \\
     \midrule
     \multirow{4}{*}{Front-back} 
     & $\checkmark$ & & & 2.13 & 6.73 & 8.71 & 10.49 \\
     & & $\checkmark$ & & 1.69 & 3.89 & 4.67 & 5.69 \\
     & & & $\checkmark$ & \textbf{14.61} & \textbf{63.97} & \textbf{77.91} & \textbf{88.85} \\
     & $\checkmark$ & $\checkmark$ & $\checkmark$ & \textbf{14.61} & \textbf{63.97} & \textbf{77.91} & \textbf{88.85} \\
    \hline
\end{tabular}
\label{tab.EEE}
\end{table}

In our proposed method, three energy functions are simultaneously optimized to obtain the optimal birotation solution. To validate the necessity of this strategy, we conduct an ablation study on the contribution of each energy function. As shown in Table~\ref{tab.EEE}, optimizing a single energy function achieves performance comparable to optimizing all three energy functions when the ground-truth relative poses are closely aligned with the corresponding basis transformation. However, its performance degrades significantly in the other two cases, demonstrating the practical necessity of optimizing all three energy functions.

\begin{table}[t!]
\settablefont
\centering
\caption{Ablation study on constraint adjustments.}
\begin{tabular}{cccccc}
    \toprule
    {Dataset} & {Adjustments} & @$1^{\circ}$ & @$3^{\circ}$ & @$5^{\circ}$ & @$10^{\circ}$\\    
    \midrule
     \multirow{2}{*}{ScanNet \cite{dai2017scannet}} 
     &  & 0.97 &  9.17 & 17.28 & 31.03 \\
     & $\checkmark$ & \textbf{1.08} & \textbf{9.56} & \textbf{18.65} & \textbf{33.08} \\
     \midrule
     \multirow{2}{*}{YFCC100M \cite{thomee2016yfcc100m}} 
     & & 9.49 &   29.73 & 42.61 &  59.46 \\
     & $\checkmark$ & \textbf{10.15} & \textbf{32.02} & \textbf{45.89} & \textbf{63.45} \\
     \hline
\end{tabular}
\label{tab.uneven}
\end{table}

\begin{table*}[t!]
\settablefont
\centering
\caption{Comparisons of different initialization strategies on the ScanNet and YFCC100M datasets.}
\begin{tabular*}{0.9\textwidth}{@{\extracolsep\fill}c r c c c c c c c c}
    \toprule
    \multirow{2}{*}{Dataset} & \multirow{2}{*}{Algorithm} & \multicolumn{4}{c}{Dense Correspondence} & \multicolumn{4}{c}{Sparse Correspondence} \\
    \cmidrule{3-6} \cmidrule{7-10}
     & & @$1^{\circ}$ & @$3^{\circ}$ & @$5^{\circ}$ & @$10^{\circ}$ & @$1^{\circ}$ & @$3^{\circ}$ & @$5^{\circ}$ & @$10^{\circ}$ \\
    \midrule
    \multirow{6}{*}{ScanNet} 
        & Nist{\'e}r \cite{nister2004efficient} & 0.56 & 6.84 & 15.24 & 31.49 & 0.74 & 7.82 & 16.54 & 33.44 \\
        & \textbf{+Ours} & \textbf{1.08} & \textbf{9.56} & \textbf{18.65} & \textbf{33.08} & \textbf{1.05} & \textbf{8.53} & \textbf{17.75} & \textbf{33.94} \\
        \cmidrule{2-10}
        & Hartley \cite{hartley1997defense} & 0.50 & 7.50 & 16.28 & 32.71 & 0.55 & 6.92 & 15.38 & 32.14 \\
        & \textbf{+Ours} & \textbf{0.98} & \textbf{9.22} & \textbf{18.12} & \textbf{32.74} & \textbf{1.09} & \textbf{8.65} & \textbf{17.69} & \textbf{33.57} \\
        \cmidrule{2-10}
        & Barath {\etal} \cite{barath2019magsac} & 0.89 & 9.30 & 18.26 & \textbf{34.41} & 0.66 & 8.01 & 17.15 & \textbf{35.11} \\
        & \textbf{+Ours} & \textbf{1.11} & \textbf{9.59} & \textbf{18.86} & 34.07 & \textbf{1.08} & \textbf{8.98} & \textbf{18.23} & 34.46 \\
    \midrule
    \multirow{6}{*}{YFCC100M} 
        & Nist{\'e}r \cite{nister2004efficient} & 6.57 & 26.84 & 41.65 & 62.01 & 6.95 & 26.74 & 41.18 & \textbf{60.74} \\
        & \textbf{+Ours} & \textbf{10.15} & \textbf{32.02} & \textbf{45.89} & \textbf{63.45} & \textbf{9.49} & \textbf{29.19} & \textbf{41.67} & 58.22 \\
        \cmidrule{2-10}
        & Hartley \cite{hartley1997defense} & 8.60 & 30.16 & 44.83 & \textbf{64.05} & 8.04 & 28.24 & \textbf{42.31} & \textbf{61.54} \\
        & \textbf{+Ours} & \textbf{10.14} & \textbf{32.06} & \textbf{45.83} & 63.23 & \textbf{9.61} & \textbf{29.35} & 41.93 & 58.41 \\
        \cmidrule{2-10}
        & Barath {\etal} \cite{barath2019magsac} & 9.36 & 33.17 & \textbf{48.53} & \textbf{67.40} & 9.44 & \textbf{32.32} & \textbf{46.48} & \textbf{64.34} \\
        & \textbf{+Ours} & \textbf{10.60} & \textbf{33.22} & 47.40 & 64.88 & \textbf{9.89} & 30.27 & 42.90 & 59.39 \\
    \hline
\end{tabular*}
\label{tab.init_comparison}
\end{table*}

\begin{figure*}[t]
    \centering
    \includegraphics[width=0.99\textwidth]{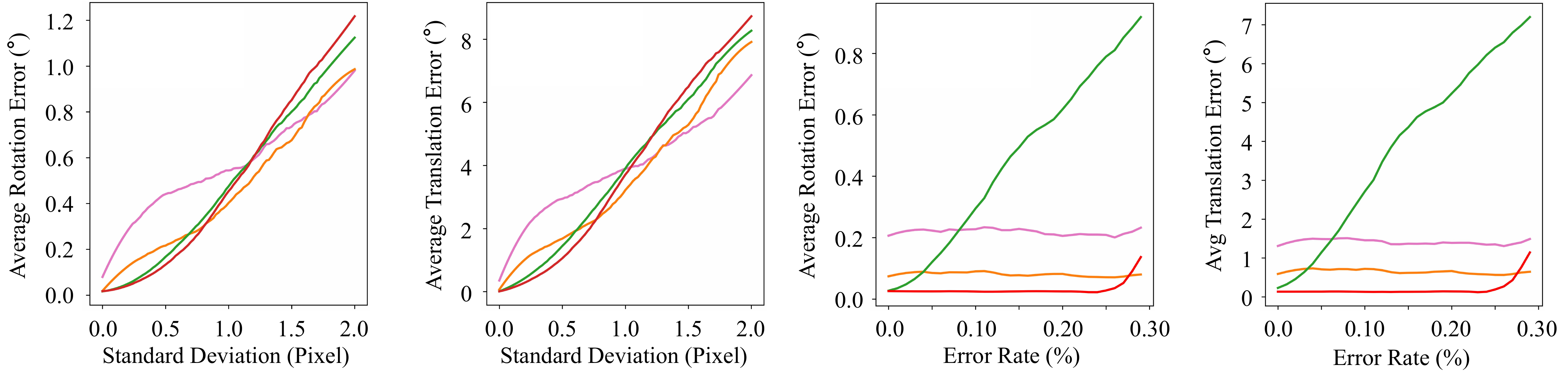}
    \caption{Average rotation and translation errors with respect to the correspondence noise levels and matching error rates. \protect\tikz[baseline=-0.5ex]{\protect\draw[MyPink, thick] (0,0) -- (0.35,0);} Nist{\'e}r \cite{nister2004efficient}, \protect\tikz[baseline=-0.5ex]{\protect\draw[MyOrange, thick] (0,0) -- (0.35,0);} Ling {\etal} \cite{ling2016high}, \protect\tikz[baseline=-0.5ex]{\protect\draw[MyGreen, thick] (0,0) -- (0.35,0);} Zhao {\etal} \cite{zhao2024dive}, and \protect\tikz[baseline=-0.5ex]{\protect\draw[red, thick] (0,0) -- (0.35,0);} ours.}
    \label{fig.noise}
\end{figure*}

In addition, to demonstrates the superiority of the adjusted constraints shown in \eqref{eq.new_constraint_1}-\eqref{eq.new_constraint_3} for residual accumulation, we conduct another ablation study on the ScanNet and YFCC100M datasets, both of which exhibit high diversity in ground-truth relative poses. As shown in Table~\ref{tab.uneven}, the adjusted constraints enable our algorithm to increase the average AUC by approximately $8\%$ on the ScanNet dataset and around $7\%$ on the YFCC100M dataset across different thresholds.

\subsection{Robustness Analyses}
\label{sec.Virtual_Experiments}

We first evaluate the performance of our algorithm with respect to different initialization strategies on the ScanNet and YFCC100M datasets. In addition to the default method \cite{nister2004efficient}, we also use two other explicit pose estimation algorithms: the direct linear transformation approach \cite{hartley1997defense} and the five-point algorithm combined with marginalizing sample consensus (MAGSAC) \cite{barath2019magsac}. As shown in Table~\ref{tab.init_comparison}, our method performs stably across different initial relative poses on both datasets. Moreover, it achieves significant improvements over the methods used for initialization on the ScanNet dataset. However, on the YFCC100M dataset, our method performs slightly worse than the algorithm \cite{barath2019magsac}, as MAGSAC is particularly effective at removing a large number of outliers.

We also compute the average rotation and translation errors of the four top-performing algorithms on the two synthetic datasets to evaluate the robustness of our method under varying correspondence noise levels and matching error rates, respectively. As illustrated in Fig.~\ref{fig.noise}(a)–(b), when the standard deviation of Gaussian noise on correspondences is below 0.8 pixels, our algorithm outperforms existing approaches. Nonetheless, as the standard deviation exceeds 0.8 pixels, the average rotation and translation errors of our method increase significantly compared to others. Additionally, as shown in Fig. \ref{fig.noise}(c)-(d), our method consistently achieves the lowest rotation and translation errors compared to existing approaches when the correspondence matching error rate is below 0.28. Nevertheless, as the proportion of incorrect correspondences increases, the accuracy of our method gradually degrades. These results indicate that our method is highly effective in scenarios with relatively accurate correspondences, but its performance degrades under conditions of extremely low-quality correspondence matching.

\section{Discussion}
\label{sec.discussion}
As mentioned earlier in the stereo camera extrinsic online calibration experiment, recovering the relative pose from a birotation solution is non-unique, resulting in outliers in the estimated relative poses. This section provides detailed analyses of this issue.

We begin by revisiting the non-uniqueness in essential matrix decomposition. Four possible solutions exist because the epipolar constraint remains satisfied when one or both camera coordinate systems are rotated by $\pi$ radians around the translation vector. Notably, when the relative pose aligns with one of the basis transformations, the epipolar constraint \eqref{eq.plEpr} simplifies to the corresponding constraint in \eqref{eq.constraints_1}, \eqref{eq.constraints_2}, and \eqref{eq.constraints_3}. Therefore, similar to the essential matrix decomposition, recovering the relative pose from a birotation solution also yields four possible solutions. Specifically, the constraint in \eqref{eq.constraints_1} can be derived from the following two transformations:
\begin{align}
    \boldsymbol{p}^{C}_1 &= \boldsymbol{p}^{C}_2 + s_1\boldsymbol{l}_1, \\
    \boldsymbol{p}^{C}_1 &= \boldsymbol{R}_x(\pi){\boldsymbol{p}^{C}_2} + s_1\boldsymbol{l}_1,
\end{align}
where 
\begin{equation}
    \boldsymbol{R}_x(\pi) = \begin{pmatrix}1 & 0 & 0 \\ 0 & -1 & 0 \\ 0 & 0 & -1 \end{pmatrix}
\end{equation}
denotes a $\pi$-radian rotation around the X-axis. The first birotation solution thus corresponds to four possible relative poses expressed as follows:
\begin{align}
    \boldsymbol{R} &= \boldsymbol{R}^{\top}_2\boldsymbol{R}_1, \ \ \boldsymbol{t} = -|s_1|\boldsymbol{R}^{\top}_2\boldsymbol{l}_1, \\
    \boldsymbol{R} &= \boldsymbol{R}^{\top}_2\boldsymbol{R}_1, \ \ \boldsymbol{t} = |s_1|\boldsymbol{R}^{\top}_2\boldsymbol{l}_1, \\
    \boldsymbol{R} &= \boldsymbol{R}^{\top}_2\boldsymbol{R}_x(\pi)\boldsymbol{R}_1, \ \ \boldsymbol{t} = -|s_1|\boldsymbol{R}^{\top}_2\boldsymbol{l}_1, \\
    \boldsymbol{R} &= \boldsymbol{R}^{\top}_2\boldsymbol{R}_x(\pi)\boldsymbol{R}_1, \ \ \boldsymbol{t} = |s_1|\boldsymbol{R}^{\top}_2\boldsymbol{l}_1. 
\end{align}
Similarly, the constraint in \eqref{eq.constraints_2} can also be derived from the following two transformations:
\begin{align}
    \boldsymbol{p}^{C}_1 &= \boldsymbol{p}^{C}_2 + s_2\boldsymbol{l}_2, \\
    \boldsymbol{p}^{C}_1 &= \boldsymbol{R}_y(\pi){\boldsymbol{p}^{C}_2} + s_2\boldsymbol{l}_2,
\end{align}    
where 
\begin{equation}
    \boldsymbol{R}_y(\pi) = \begin{pmatrix} -1 & 0 & 0 \\ 0 & 1 & 0 \\ 0 & 0 & -1 \end{pmatrix}
\end{equation}
denotes a $\pi$-radian rotation around the Y-axis. The second birotation solution corresponds to four possible relative poses expressed as follows:
\begin{align}
    \boldsymbol{R} &= \boldsymbol{R}^{\top}_2\boldsymbol{R}_1, \ \ \boldsymbol{t} = -|s_2|\boldsymbol{R}^{\top}_2\boldsymbol{l}_2, \\
    \boldsymbol{R} &= \boldsymbol{R}^{\top}_2\boldsymbol{R}_1, \ \ \boldsymbol{t} = |s_2|\boldsymbol{R}^{\top}_2\boldsymbol{l}_2, \\
    \boldsymbol{R} &= \boldsymbol{R}^{\top}_2\boldsymbol{R}_y(\pi)\boldsymbol{R}_1, \ \ \boldsymbol{t} = -|s_2|\boldsymbol{R}^{\top}_2\boldsymbol{l}_2, \\
    \boldsymbol{R} &= \boldsymbol{R}^{\top}_2\boldsymbol{R}_y(\pi)\boldsymbol{R}_1, \ \ \boldsymbol{t} = |s_2|\boldsymbol{R}^{\top}_2\boldsymbol{l}_2. 
\end{align}
Additionally, the constraint represented in \eqref{eq.constraints_3} can be derived from the following two transformations:
\begin{align}
    \boldsymbol{p}^{C}_1 &= {\boldsymbol{p}^{C}_2} + s_3\boldsymbol{l}_3, \\
    \boldsymbol{p}^{C}_1 &= \boldsymbol{R}_z(\pi){\boldsymbol{p}^{C}_2} + s_3\boldsymbol{l}_3,
\end{align}
where 
\begin{equation}
\boldsymbol{R}_z(\pi) = \begin{pmatrix} -1 & 0 & 0 \\ 0 & -1 & 0 \\ 0 & 0 & 1 \end{pmatrix}    
\end{equation}
denotes a $\pi$-radian rotation around the Z-axis. The third birotation solution thus corresponds to four possible relative poses expressed as follows:
\begin{align}
    \boldsymbol{R} &= \boldsymbol{R}^{\top}_2\boldsymbol{R}_1, \ \ \boldsymbol{t} = -|s_3|\boldsymbol{R}^{\top}_2\boldsymbol{l}_3, \\
    \boldsymbol{R} &= \boldsymbol{R}^{\top}_2\boldsymbol{R}_1, \ \ \boldsymbol{t} = |s_3|\boldsymbol{R}^{\top}_2\boldsymbol{l}_3, \\
    \boldsymbol{R} &= \boldsymbol{R}^{\top}_2\boldsymbol{R}_z(\pi)\boldsymbol{R}_1, \ \ \boldsymbol{t} = -|s_3|\boldsymbol{R}^{\top}_2\boldsymbol{l}_3, \\
    \boldsymbol{R} &= \boldsymbol{R}^{\top}_2\boldsymbol{R}_z(\pi)\boldsymbol{R}_1, \ \ \boldsymbol{t} = |s_3|\boldsymbol{R}^{\top}_2\boldsymbol{l}_3. 
\end{align}
Similar to the process of determining the correct relative pose after essential matrix decomposition, all possible relative poses from birotation solutions can be enumerated, and the correct one can be identified by checking the depth signs of the reconstructed 3-D points. Nonetheless, in our proposed relative pose estimation algorithm, this process can be omitted when performing appropriate initialization. When the initial relative pose obtained from essential matrix decomposition is close to the correct solution, the regularization term in \eqref{eq.E} ensures that the birotation solution does not converge to incorrect ones. Although our approach consistently produces unique solutions, as demonstrated in our experiments, there remains a potential risk of yielding non-unique solutions in practical applications, which we leave for future work.

\section{Conclusion and Future Work}
\label{sec.Conclusion}

This article introduced a novel birotation solution for general relative pose problems. To expand the applicability of existing birotation-based methods, this study first presented three basis transformations and the corresponding geometric metrics, which serve as the foundation for the formulation of three energy functions. Relative poses can be accurately and robustly estimated by simultaneously minimizing these energy functions.
This high-performing relative pose estimation algorithm significantly overcame the limitations of existing birotation-based methods, particularly by reducing reliance on initialization quality and alleviating the non-uniqueness in relative pose recovery. Extensive experiments conducted on both real-world and synthetic datasets demonstrate the superior performance of our algorithm over existing relative pose estimation methods across varying correspondence densities, as well as its strong robustness to different correspondence noise levels and matching error rates.

In the future, we are committed to further advancing our approach along the following three directions: (1) improving its efficiency by avoiding the simultaneous optimization of all three energy functions, (2) enhancing its robustness by incorporating a self-evolving correspondence matching network, and (3) ensuring the uniqueness of relative pose recovery from a birotation solution.

\bibliographystyle{IEEEtran}
\bibliography{main}

\end{document}